\documentclass[onefignum,onetabnum]{siamonline220329}

\usepackage{lipsum}
\usepackage{subfig}
\usepackage{multirow}
\usepackage{amsfonts, amssymb}
\usepackage{graphicx}
\usepackage{epstopdf}
\usepackage{algorithmic}
\ifpdf
  \DeclareGraphicsExtensions{.eps,.pdf,.png,.jpg}
\else
  \DeclareGraphicsExtensions{.eps}
\fi

\usepackage{enumitem}
\setlist[enumerate]{leftmargin=.5in}
\setlist[itemize]{leftmargin=.5in}

\newsiamremark{remark}{Remark}
\newsiamremark{hypothesis}{Hypothesis}
\crefname{hypothesis}{Hypothesis}{Hypotheses}
\newsiamthm{claim}{Claim}

\headers{Approximation by Wasserstein GAN}{Y.Gao, M. Ng, M. Zhou}

\title{Approximating Probability Distributions by using Wasserstein Generative Adversarial Networks \thanks{Accepted by SIAM Journal on Mathematics of Data Science (SIMODS).
\funding{This work was funded by 
HKRGC GRF 12300519, 17201020 and 17300021,
HKRGC CRF C1013-21GF and C7004-21GF,
and Joint NSFC and RGC N-HKU769/21.
}}}

\author{Yihang Gao\thanks{Department of Mathematics, The University of Hong Kong, Pokfulam, Hong Kong
  (\email{gaoyh@connect.hku.hk}).}
\and 
Michael K. Ng \thanks{Department of Mathematics, The University of Hong Kong, Pokfulam, Hong Kong
  (\email{mng@maths.hku.hk}).}
\and
Mingjie Zhou\thanks{Department of Mathematics, The University of Hong Kong, Pokfulam, Hong Kong
  (\email{mjzhou@connect.hku.hk}).}}
\usepackage{amsopn}

\ifpdf
\hypersetup{
  pdftitle={Approximating Probability Distributions by using Wasserstein Generative Adversarial Networks},
  pdfauthor={Y. Gao, M. Ng, M. Zhou}
}
\fi

\begin{document}

\maketitle

% REQUIRED
\begin{abstract} 
Studied here are Wasserstein generative adversarial networks (WGANs) with GroupSort neural networks as their discriminators.
It is shown that the error bound of the approximation for the target distribution depends on the width and depth (capacity) of the generators and discriminators and the number of samples in training. A quantified generalization bound is established for the Wasserstein distance between the generated and target distributions. 
According to the theoretical results, WGANs have a higher requirement for the capacity of discriminators than that of generators, which is consistent with some existing results. More importantly, the results with overly deep and wide (high-capacity) generators may be worse  than those with low-capacity generators if discriminators are insufficiently strong. 
Numerical results obtained using Swiss roll and MNIST datasets confirm the theoretical results.
%By using machine learning techniques, 
%we further develop a generalization error bound, which is free from the curse of %dimensionality
%with respect to 
%numbers of training data:
%. It reduces from 
%$\widetilde{\mathcal{O}}(m^{-1/r} + n^{-1/d})$ to $\widetilde{\mathcal{O}}
%(\text{Pdim}(\mathcal{F}\circ \mathcal{G}) \cdot m^{-1/2} + \text{Pdim}
%(\mathcal{F}) \cdot n^{-1/2})$,
%where $m$ and $n$ are the numbers of training samples
%for a generator $\mathcal{G}$
%and a discriminator $\mathcal{F}$ respectively, 
%$r$ and $d$ are the input and output dimensions respectively, and $\text{Pdim}%%(\cdot)$ denotes the pseudo-dimension of neural networks.
%However, we observed an interesting phenomenon between the %derived error %bounds and intuition.}
%In contrast to with existing results, our results are %applicable to general %GroupSort WGAN without special %architectures, \textcolor{purple}{thanks to the %approximation capabilities of GroupSort neural networks, %where injective or %invertible neural networks are required %in some prior works}. 
%Finally, numerical results on the swiss roll and MNIST data %sets confirm our %theoretical results.
\end{abstract}

% REQUIRED

% REQUIRED

\section{Introduction}
First proposed by \cite{goodfellow2014generative}, generative adversarial networks (GANs) have become popular in recent years for their remarkable abilities to approximate various types of distributions, e.g.,  probability distributions in statistics, and distributions for image pixels \cite{goodfellow2014generative, odena2017conditional} and natural language \cite{bowman2015generating}. However, although the success of GANs has been shown empirically, they remain insufficiently understood theoretically.  Even if the discriminator is fooled and the generator wins the game in training, we usually cannot determine whether the generated distribution $\mathcal{D}$ is sufficiently close to the target distribution 
${\cal D}_{target}$. Arora et al. \cite{arora2017generalization} found that if the discriminator is insufficiently strong  (with a fixed architecture and constrained parameters), then $\mathcal{D}$ can be far away from $\mathcal{D}_{target}$, even though the learning has great generalization properties (i.e., small empirical loss). Bai et al. \cite{bai2018approximability} designed some special discriminators with restricted approximability to let the trained generator approximate the target distribution in statistical metrics, i.e., the Wasserstein distance. However, their discriminator design  is not applicable because it pertains only to some specific statistical distributions (e.g., exponential families), and it requires using
invertible or injective generators, i.e., it  requires implementing training weights under a special setting.  
Liang \cite{liang2021well} proposed a type of oracle inequality to develop the generalization error of GANs' learning in total variation (TV) distance. Several assumptions (e.g., specially designed generators and discriminators, and the realization of generators for the target distribution) are required, and the Kullback–Leibler (KL) divergence is used to measure the distance between distributions. However, the KL divergence may be inappropriate for the generalization (convergence) analysis of GANs because many natural distributions lie on low-dimensional manifolds. The support of the generated distribution must coincide well with that of the target distribution, otherwise, the KL divergence between the
generated and target distributions 
is infinite, which may not be helpful in generalization analysis. After all, theoretical error bounds are usually much stricter than the exact errors in practice. Bai et al. \cite{bai2018approximability} and Liang \cite{liang2021well} showed the effectiveness of GANs in approximating probability distributions to some extent, although their results are for some special neural network (NN) architectures.
\cref{tab:list_methods} summarizes the above results for GANs.
%understanding.

\begin{table}[h]
\caption{Comparisons of some relevant results.
Here, 
$m$ and $n$ are the numbers of training samples
for a generator $\mathcal{G}$
and a discriminator $\mathcal{F}$, respectively, 
$r$ and $d$ are the input and output dimensions, respectively, and $\text{Pdim}(\cdot)$ denotes the pseudo-dimension of neural networks (NNs).
} 
%Note that in the first two results, constants involved in upper %bounds are related to neural network size.}
\label{tab:list_methods}
\footnotesize
\centering
\begin{tabular}{| c|c|c |} 
\hline
Method & Metric & Upper bound \\
\hline
Arora et al. \cite{arora2017generalization} & Neural net distance & $\widetilde{\mathcal{O}}\left(m^{-1/2}+n^{-1/2}\right)$\\
& & (here, the constants involved are related to \\
& & the NN size.)\\
\hline
Bai et al. \cite{bai2018approximability} & 
Wasserstein distance & 
$\widetilde{\mathcal{O}}\left(m^{-1/2}+n^{-1/2}\right)$\\
(special requirements for
& & (here, the constants involved are related to \\
discriminator and generator 
& & the NN size.) \\
architecture) & & \\
\hline
Liang \cite{liang2021well} & TV distance & 
$\widetilde{\mathcal{O}}\left(\sqrt{\text{Pdim}(\mathcal{F}\circ \mathcal{G}) }
 \cdot \left(m^{-1/2}+n^{-1/2}\right)\right)$\\
(special requirements for
& & \\
discriminator and generator && \\
architecture) & & \\
\hline
Proposed analysis & Wasserstein distance 
& $\min \left\{\widetilde{\mathcal{O}}\left(\sqrt{\text{Pdim}(\mathcal{F}\circ \mathcal{G})/m}+ \sqrt{\text{Pdim}(\mathcal{F})/n } \right), \right.$ \\
(GroupSort discriminator) 
& & $\left.\widetilde{\mathcal{O}}\left(m^{-1/r} + n^{-1/d}\right)\right\}$ \\
\hline
\end{tabular}
\end{table}

NNs have proven extraordinary expressive power. Sufficiently wide two-layer NNs can approximate any continuous function on compact sets \cite{hornik1991approximation, cybenko1989approximation, luo2020two}, and deep NNs are capable of approximating wider ranges of functions \cite{yarotsky2017error, yarotsky2018optimal}. 
After applying deep generative models (e.g., variational autoencoders and GANs) widely, researchers began to study the capacity of deep generative networks for approximating distributions. Recently, Yang et al. \cite{yang2022capacity} showed that sufficiently wide and deep NNs can generate a distribution that is arbitrarily close to a high-dimensional target distribution in Wasserstein distance by inputting only a one-dimensional source distribution (a similar conclusion is obtained for multi-dimensional source distributions). 

However, although deep generative networks can approximate distributions, discriminators play a crucial role in GANs. If the discriminator is insufficiently strong (e.g. ReLU NNs with fixed architectures and constrained parameters), then it is easily fooled by an imperfect generator. Fortunately, it has been shown both empirically and theoretically that GroupSort NNs \cite{tanielian2021approximating, anil2019sorting}, which are stronger than ReLU NNs, can approximate any 1-Lipschitz function arbitrarily close in a compact set with sufficient width and depth. Therefore, in our generalization analysis of WGANs, we use GroupSort NNs as discriminators.

\subsection{Our contributions}

Motivated by the aforementioned  research, herein we study the approximation of distributions by using (GroupSort) WGANs, and our main contributions are as follows.

%\noindent
%\begin{enumerate}
%\item[(1)] 

(i) We conduct the generalization analysis for WGANs with nonparametric discriminators (see \cref{thm_nonpara_bound}). Next, we analyze 
%, however, the setting is not applicable and accessible in real %applications. It motivates us to analyze 
WGANs with GroupSort NNs as discriminators.
We connect the capacity of deep generative networks and GroupSort networks to derive a generalization bound for GroupSort WGANs with finite samples (see \cref{main_thm}), i.e.,
$\mathcal{O}(m^{-1/r} + n^{-1/d})$, 
where $m$ and $n$ are the numbers of training samples
for generators in $\mathcal{G}$
and discriminators in $\mathcal{F}$, respectively, and 
$r$ and $d$ are the input and output dimensions, respectively.
This implies that with sufficiently many training samples, a WGAN with properly chosen width and depth for generators and discriminators can approximate distributions well. 
To the best of our knowledge, this is the first paper to derive the generalization bound in Wasserstein distance for WGANs in terms of the depth and width of the generators (ReLU NNs) and discriminators (GroupSort NNs), as well as the number of training samples. Compared with existing results \cite{bai2018approximability, liang2021well}, our results are suitable for 
GroupSort WGANs without special requirements for the NN architecture. 

%\noindent
(ii) 
%\item[(2)] 
We analyze the importance of both generators and discriminators in WGANs (see discussions in \cref{remark1}, \cref{remark2} and experimental results in \cref{syn_g_d} and \cref{mnist_g_d}). We 
show that for nonparametric discriminators (that are sufficiently strong; see \cref{thm_nonpara_bound}), the generalization is easier to be obtained, but WGANs with GroupSort NNs (as discriminators) have high requirements for the capability of discriminators according to our derived generalization bound (see \cref{main_thm}). In other words, the results with overly deep and wide (high-capacity) generators (without regularization) may be worse than those with low-capacity generators if the discriminators are insufficiently strong.
%\noindent

(iii)
To overcome the undesired curse of dimensionality,
we develop another version of the generalization error bound based on techniques in
Rademacher complexity (see \cref{new_generalization_thm}). 
%Interestingly, although 
The new derived error bound is of $\widetilde{\mathcal{O}}\left(\sqrt{\text{Pdim}(\mathcal{F}\circ \mathcal{G})} \cdot m^{-1/2} + \sqrt{\text{Pdim}(\mathcal{F})} \cdot n^{-1/2}\right)$, where $\text{Pdim}(\cdot)$ denotes the pseudo-dimension of NNs. 

%, it contradicts to numerical observations that the capacity of %discriminators contributes to lower generalization error.   
The rest of the paper is as follows. In Section 2,
%\cref{prelim}, 
we give notation and mathematical preliminaries. In 
Section 3,
%\cref{analysis}, 
we study generalization analysis for GroupSort WGANs (by two different techniques) and show the above theoretical results. In Section 4, we conduct numerical experiments to verify our theoretical results. Finally, we make some concluding remarks in Section 5.
%\cref{conclusion}.

\section{Background and preliminaries}
\label{prelim}
\subsection{Notation}

Throughout the paper, we use $\nu$, $\pi$, and $\mu_{\theta}$ to denote the target distribution, the source distribution (as an input to the generator), and the generated distribution (with $\theta$ as the set of parameters for the generator), respectively. More concretely, $\mu_{\theta}$ is the distribution of $g_{\theta}(Z)$, where $Z \sim \pi$ and $g_{\theta}(\cdot)$ is the generator. The source distribution $\pi$ lies on $\mathbb{R}^r$, and both the target and the generated distributions are defined on $\mathbb{R}^d$, where $r$ and $d$ are the numbers of dimensions. 
In Section 3,
%\cref{analysis}, 
we use $m$ for the number of generated samples from $\mu_{\theta}$ (the same number for samples from the source distribution $\pi$) and $n$ for the number of real data from the target distribution $\nu$. The symbol $\lesssim$ is used to hide some unimportant constant factors. For two functions $\ell$ and $h$ of a number $n$, $\ell \lesssim h$ indicates $\ell \leq C \cdot h$ for a non-negative constant C, and equivalently, $\ell = \mathcal{O}(h)$.
We adopt $\widetilde{\mathcal{O}}(\cdot)$ to omit poly-logarithmic terms. For two positive real numbers or functions $A$ and $B$, the term $A \vee B $ is equivalent to $\max \{A,B\}$ and $A \wedge B$ is equivalent to $\min\{A,B\}$. For a matrix $\mathbf{V}=(\mathbf{V}_{i,j})$ of size $p \times q$, we let $||\mathbf{V}||_{\infty}=\sup_{||x||_{\infty}=1}||\mathbf{V}\mathbf{x}||_{\infty}$. We may also use the $(2,\infty)$ norm of $\mathbf{V}$, i.e. $||\mathbf{V}||_{2, \infty}=\sup_{||x||_{2}=1}||\mathbf{V}\mathbf{x}||_{\infty}$.

\subsection{Wasserstein distance and Wasserstein GANs}
\label{w_distance_wgan}

Several types of probability metrics are available for measuring the distance between two distributions. As noted 
in Section 1, the $\text{KL}$ divergence (and its special case the Jensen–Shannon divergence) is  unsuitable for the generalization error analysis of GANs because of its non-convergence; see \cite{arjovsky2017wasserstein} for a detailed discussion.
Instead, the Wasserstein distance is a more relaxed and flexible metric and can be generalized with some mild conditions (e.g., finite 3-moment for the distribution). In our analysis, we use the Wasserstein-1 distance ($\mathbf{Wass}_1$) as the evaluation metric between the generated and the target distributions. The Wasserstein-$p$ distance ($p \geq 1$) between two distributions $\mu$ and $\nu$ is defined as
$$
\mathbf{Wass}_p(\mu,\nu) = \inf_{\gamma \in \Gamma(\mu,\nu)} \int ||x-y||^p d(\gamma(x,y)),
$$
where $\Gamma(\mu,\nu)$ denotes the set of all joint distributions $\gamma(x,y)$ whose marginals are $\mu$ and $\nu$ respectively. When $p=1$, the Kantorovich-Rubinstein duality \cite{villani2008optimal} gives an alternative definition of $\mathbf{Wass}_1$:
\begin{equation}
%\begin{aligned}
        \mathbf{Wass}_1(\mu,\nu)
				%& 
				=\sup_{||f||_{\text{Lip}} \leq 1} \left \arrowvert  \mathbb{E}_{X\sim \mu} f(X)- \mathbb{E}_{X\sim \nu} f(X)\right  \arrowvert\\
        %& 
				= \sup_{||f||_{\text{Lip}}\leq 1} \left\{\mathbb{E}_{X\sim \mu} f(X)- \mathbb{E}_{X\sim \nu} f(X) \right\},
%\end{aligned}
\end{equation}
where the last equation holds because of the symmetry ($||f||_{\text{Lip}}=||-f||_{\text{Lip}}$).

Unlike the original model in \cite{goodfellow2014generative}, GANs are usually formulated in a more general form:
\begin{equation}
\label{general_GAN}
    \min_{g_{\theta} \in \mathcal{G}} \max_{f \in \mathcal{F}} \left\{ \mathbb{E}_{Z \sim \pi} f(g_{\theta}(Z)) - \mathbb{E}_{X \in \nu} f(X) \right\},
\end{equation}
where $\mathcal{G}$ and $\mathcal{F}$ are the generator and the discriminator classes, respectively. Here, we assume that our pre-defined discriminator class is symmetric, i.e., $f \in \mathcal{F}$ and $-f \in \mathcal{F}$. The symmetry condition is easily satisfied, 
e.g., feed-forward NNs without activation functions in the last layer. Intuitively, the capacity of generator class $\mathcal{G}$ determines the highest quality of generated distributions, and the discriminator class $\mathcal{F}$ guides the learning process, pushing the learned parameters closer to the optimal ones. In practice, we can only access finitely many numbers of samples from the target distribution and limited by computing power, we deal with 
finitely many generated data.
% is considered due to limited computing power. 
Empirically, instead of \cref{general_GAN}, we must solve the following adversarial problem and obtain estimated parameters for the generator
as follows:
\begin{equation}
\label{minmax}
    \hat{\theta}_{m,n} = \arg \min_{g_{\theta} \in \mathcal{G}} \max_{f \in \mathcal{F}} \{\hat{\mathbb{E}}^m_{Z \sim \pi} f(g_{\theta}(Z))-\hat{\mathbb{E}}^n_{X \sim \nu} f(X)\},
\end{equation}
where $\hat{\mathbb{E}}^m_{Z \sim \pi}$ (resp. $\hat{\mathbb{E}}^n_{X \sim \nu}$) denotes the expectation on the empirical distribution of $m$ (resp. $n$) i.i.d. samples of random variables $Z$ (resp. $X$) from $\pi$ (resp. $\nu$). Therefore, the output generator $g_{\hat{\theta}_{m,n}}$ is obtained by solving \cref{minmax}.

Suppose that we have the following generator class $\mathcal{G}$:
\begin{equation}
    \label{generator_class}
    \mathcal{G}=\{g_{\theta}(\mathbf{x})= \text{NN}(\mathbf{x}|W_g,D_g,\theta): g_{\theta}(\mathbf{x}) \text{ is L-Lipschitz and } \|g_{\theta}(\mathbf{0})\|_2 \leq M,  \theta=\{\mathbf{W}_i,\mathbf{b}_i\}_{i=1}^{D_g} \},
\end{equation}
where $\text{NN}(\mathbf{x}|W_g,D_g,\theta)$ denotes NNs with $D_g$ layers (depth), $W_g$ neurons in each layer (width), and $\theta$ as the set of parameters. Here, we require the NNs to be $L$-Lipschitz and with bounded bias, which are very common in practice. Actually, some of the numerical tricks, e.g., $L_2$ regularization and data (or batch) normalization that stabilize the training, account for the reasonableness of Lipschitzness and boundedness requirements. We use $\mathbf{W}_i$ to represent the matrix of linear transform between layers $i-1$ and $i$. Here, we use ReLU as the activation function $\sigma$. The architecture of the NN is formulated as follows:
$$
\text{NN}(\mathbf{x}|W_g,D_g,\theta)=\mathbf{W}_{D_g} \cdot \sigma(\mathbf{W}_{D_g-1}\cdot \sigma( \cdots \sigma(\mathbf{W}_1 \cdot \mathbf{x}+\mathbf{b}_1)+ \cdots) +\mathbf{b}_{D_g-1})+\mathbf{b}_{D_g}.
$$

\subsection{GroupSort neural networks}
% and GroupSort Wassertein GAN}

For Wasserstein GANs \cite{arjovsky2017wasserstein}, the discriminator 
class $\mathcal{F}_W$ must be 
\begin{equation}
\label{W_discriminator_class}
    \mathcal{F}_W=\{f: ||f||_{\text{Lip}} \leq 1\},
\end{equation}
where $||f||_{\text{Lip}}$ denotes the Lipschitz constant of $f$. In Section 3.1,
%\cref{generalization_nonpara}, 
we analyze the generalization error bound of WGANs with $\mathcal{F}_W$ as the discriminator class. However, the discriminator class $\mathcal{F}_W$ is empirically impractical, and in practice, several strategies are used to approximate $\mathcal{F}_W$ in WGANs. Most studies use  ReLU (or LeakyReLU) NNs with $1$-Lipschitz constraints on parameters as a surrogate for $\mathcal{F}_W$. Note that Yarotsky \cite{yarotsky2017error} showed that ReLU NNs can approximate any 1-Lipschitz function on a compact set arbitrarily close if they are sufficiently wide and deep, where their weights cannot be constrained. In other words, the approximation capabilities of ReLU NNs to 1-Lipschitz functions are not theoretically guaranteed if their weights are constrained (as is usually done in practice). Huster et al. \cite{huster2018limitations} showed that the approximation capabilities of ReLU NNs are poor  when constrained weights are used, and such NNs cannot even represent some of the simplest functions (e.g., the absolute function).

Recently, Anil et al. \cite{anil2019sorting} and Tanielian et al. \cite{tanielian2021approximating} noted that GroupSort NNs, whose activation functions are norm-preserving, have stronger approximation capabilities for 1-Lipschitz functions compared with ReLU and LeakyReLU NNs with constrained weights. As noted in \cite{anil2019sorting}, norm-preserving activation functions play an essential role in approximating $1$-Lipschitz functions. For example, if a $1$-Lipschitz function can achieve 
the $\textbf{Wass}_1$ distance, then its corresponding 
input-output gradient must have norm $1$ almost everywhere. More intuitive and theoretical explanations of GroupSort NNs can be found in \cite[Sections 3 \& 4]{anil2019sorting}, and in \cite{zhang2022rethinking}  recently, which theoretically and empirically validates the expressive power and the robustness of GroupSort NNs.   
Moreover, their experimental results on image generation further show the capability of GroupSort WGANs. Based on this, herein we analyze the generalization properties of GroupSort WGANs both theoretically and numerically.

% \subsection{GroupSort Wassertein GAN}
A GroupSort NN \cite{anil2019sorting, tanielian2021approximating} is defined as
\begin{equation}
\label{groupsortnn}
 \text{GS}(\mathbf{x}|W_f,D_f,\alpha) = \mathbf{V}_{D_f} \cdot \sigma_{k}(\mathbf{V}_{D_f-1} \cdot \sigma_{k}(\cdots \sigma_k(\mathbf{V}_1 \cdot \mathbf{x} + \mathbf{c}_1)+ \cdots )+\mathbf{c}_{D_f-1})+\mathbf{c}_{D_f},    
\end{equation}
with constraints
\begin{equation}
\label{constraints}
\begin{aligned}
\left\|\mathbf{V}_{1}\right\|_{2, \infty} \leqslant 1 \hspace{1em} \text{ and  }& \hspace{1em} \max \left(\left\|\mathbf{V}_{2}\right\|_{\infty}, \ldots,\left\|\mathbf{V}_{D_f}\right\|_{\infty}\right) \leqslant 1,
\end{aligned}    
\end{equation}
where $\alpha=\{\mathbf{V}_i,\mathbf{c}_i\}_{i=1}^{D_f}$ denotes parameters of the GroupSort NN, $W_f$ is the width of each layer, $D_f$ is the depth (number of layers), and $\sigma_k$ is the GroupSort activation function with grouping size $k$. When applying GroupSort activation function $\sigma_{k}$, we first split $W_f$ neurons into $p$ different groups where $p=W_f/k$, and $k$ elements in each group: $G_1=\{x_1, \cdots, x_k\}, \cdots, G_{p}=\{x_{pk-(k-1)}, \cdots, x_{pk}\}$, and then sort the elements in each group in decreasing order. For example, for an input vector $\mathbf{x} = [-1,1,0,-2,3,4]^{\top}$ and GroupSort activation function $\sigma_3(\cdot)$ (i.e., $k=3$), then there are two groups $G_1 = \{-1,1,0\}$, $G_2 = \{-2,3,4\}$. Applying sorting operations to each group and then combing all elements together into a new vector, we have $\sigma_3(\mathbf{x}) = [1,0,-1,4,3,-2]^{\top}$.  Visualization of the GroupSort activation function is available in \cite[Figure 1]{anil2019sorting}. Note that the nonlinear GroupSort activation function does not change the values of input vectors but rearranges elements in each group. Therefore, it is norm-preserving. 
Herein, we set $k=2$ (which is a predefined parameter for the GroupSort activation function) and choose GroupSort NNs as the discriminator class:
\begin{eqnarray}
\label{discriminator_class}
%\begin{split}
    \mathcal{F}_{\text{GS}} & = & 
		\{f_{\alpha}(\mathbf{x})=\text{GS}(\mathbf{x}|W_f,D_f,\alpha): \text{GS}(\mathbf{x}|W_f,D_f,\alpha) \ \text{is a neural network} 
		\nonumber \\ 
		& & \quad \text{of form \cref{groupsortnn} with \cref{constraints}} \}.
\end{eqnarray}
In \cref{grouping_size}, we discuss the grouping size 
for GroupSort NNs.

In the next section, we study the quality of the obtained generator (the solution of \cref{minmax}), with respect to its $\mathbf{Wass}_1$ distance to the target distribution $\nu$  (i.e., $\mathbf{Wass}_1(\mu_{\hat{\theta}_{m,n}},\nu)$) given the pre-defined generator class $\mathcal{G}$, the discriminator class $\mathcal{F}$ and $(m,n)$ samples for training.

\section{Generalization analysis for Wasserstein GANs}
\label{analysis}

In Section 3.1,
%\cref{generalization_nonpara}, 
we analyze the generalization error bound for WGAN with nonparametric discriminators $\mathcal{F}=\mathcal{F}_{W}$. In practice, GroupSort NNs $\mathcal{F}_{\text{GS}}$ are adopted as a surrogate for $\mathcal{F}_{W}$, and the corresponding generalization analysis is conducted in 
Section 3.2.
%\cref{generalization_para}. 
Furthermore, we derive two generalization error bounds for GroupSort WGANs based on different techniques (in optimal transport and machine learning, respectively). We also 
discuss some theoretical insights and observations from the error bounds. 
%An exciting contradiction of theory and numerical results is %discovered.

\subsection{Nonparametric discriminators}
\label{generalization_nonpara}
In this subsection, we study the generalization properties of WGAN by using
nonparametric discriminators
%. Here the discriminator class $\mathcal{F}$ for \cref{minmax} %is set to be the class of all 1-Lipschitz functions, i.e., 
$\mathcal{F}=\mathcal{F}_W$.
Under \cref{assump_3Moment} and \cref{convg_W1} (see below), 
we can 
%the following assumption 
guarantee that with sufficiently many samples, the empirical distributions $\hat{\nu}$ and $\hat{\pi}$ are close (convergent) to $\nu$ and $\pi$ in $\mathbf{Wass}_1$ distance, respectively. The convergence requirements for both the source and the target distributions are essential because they are fundamental for the convergence of GANs; it is almost impossible to study the generated distribution under the metric if the empirical distributions are not convergent. 

\begin{assumption}
\label{assump_3Moment}
Both $d$-dimensional target distribution $\nu$ and  
$r$-dimensional source distribution $\pi$ have finite 3-moment, i.e. $M_3(\nu)=\mathbb{E}_{X \sim \nu}\|X\|^{3}<\infty$ and $M_3(\pi)=\mathbb{E}_{Z \sim \pi}\|Z\|^{3}<\infty$.
\end{assumption}

\begin{lemma}[Convergence in $\mathbf{Wass}_1$ distance \cite{NEURIPS2020_2000f632}]
\label{convg_W1}
Assume that the distribution $\nu$ satisfies that $M_{3}=\mathbb{E}_{X \sim \nu}|X|^{3}<\infty$, and consider $\mathcal{F}=\{f$ :
$\left.\mathbb{R}^{d} \rightarrow \mathbb{R}: ||f||_{\text{Lip}} \leq 1\right\}$. 
%Assume that the distribution $\nu$ satisfies that %$M_{3}=\mathbb{E}_{X \sim \nu}|X|^{3}<\infty .$ 
Then there exists a constant $C$ depending only on $M_{3}$ such that
$$
\mathbb{E} \mathbf{Wass}_{1}\left(\nu, \hat{\nu}^n\right) 
\leq C \cdot\left\{\begin{array}{ll}
n^{-1 / 2}, & d=1, \\
n^{-1 / 2} \log n, & d=2, \\
n^{-1 / d}, & d \geq 3.
\end{array}\right.
$$
\end{lemma}

In \cref{convg_W1},
$C$ is a constant depending only on the 3-moment of the target distribution $\nu$ (i.e., $M_3(\nu)$), and it is independent 
of the number of samples and the model architecture. Next, we present the main theorem for the WGANs with nonparametric discriminators. 

\begin{theorem}[Generalization error bound for WGAN with nonparametric discriminators.]
\label{orig_thm}
Suppose that \cref{assump_3Moment} holds for  the target distribution $\nu$ (lies on $\mathbb{R}^d$) and the source distribution $\pi$ (lies on $\mathbb{R}^r$). For GANs, the generator class $\mathcal{G}$ defined in \cref{generator_class} is $L$-Lipschitz and let the discriminator class $\mathcal{F}=\mathcal{F}_W$ (defined in \cref{W_discriminator_class}), i.e., WGANs. 
If the set of parameters $\hat{\theta}_{m,n}$ is obtained from \cref{minmax}, then the following holds,
\begin{equation}
\label{bound_final}
\begin{split}
& \mathbb{E} \mathbf{Wass}_1(\mu_{\hat{\theta}_{m,n}}, \nu)\\
\leq & \min_{g_{\theta} \in \mathcal{G}}\mathbf{Wass}_1(\mu_{\theta},\nu)+C_1\cdot L  \cdot\left\{\begin{array}{ll}
m^{-1 / 2}, & r=1, \\
m^{-1 / 2} \log m, & r=2, \\
m^{-1 / r}, & r \geq 3,
\end{array}\right. + C_2\cdot\left\{\begin{array}{ll}
n^{-1 / 2}, & d=1, \\
n^{-1 / 2} \log n, & d=2, \\
n^{-1 / d}, & d \geq 3,
\end{array}\right.  
\end{split}
\end{equation}
where constants $C_1$ and $C_2$ depends only on $M_3(\pi)$ and $M_3(\nu)$ respectively.
\end{theorem}

The proof of \cref{orig_thm} is based on some well-known inequalities and the convergence property of $\mathbf{Wass}_1$ in optimal transport. Here we begin by splitting the error into 
several terms, e.g., the capability of NNs (generators) to generate distributions, the generalization error of the source distribution, and the generalization error of the target distribution in terms of the $\textbf{Wass}_1$ distance. By applying the convergence property of the $\textbf{Wass}_1$ distance for Lipschitz continuous functions \cite{NEURIPS2020_2000f632}, 
we complete the proof.

%\subsection{Proof of \cref{orig_thm}}
%\label{proof_thm1}
\begin{proof}
For any ${\hat{\theta}_{m,n}}$ obtained from \cref{minmax}, we have
\begin{align*}
  & \mathbf{Wass}_1(\mu_{\hat{\theta}_{m,n}}, \nu) \\
  = & \sup_{||f||_{\text{Lip}}\leq 1} \{\mathbb{E}_{Z \sim \pi} f(g_{\hat{\theta}_{m,n}}(Z))-\mathbb{E}_{X \sim \nu}f(X)\} \\
  \leq & \sup_{||f||_{\text{Lip}}\leq 1} \{\mathbb{E}_{Z \sim \pi} f(g_{\hat{\theta}_{m,n}}(Z))- \hat{\mathbb{E}}^n_{X \sim \nu} f(X) \}+  \sup_{||f||_{\text{Lip}}\leq 1} \{ \hat{\mathbb{E}}^n_{X \sim \nu} f(X) -\mathbb{E}_{X \sim \nu}f(X)\} \\
  \leq & \sup_{||f||_{\text{Lip}}\leq 1} \{\mathbb{E}_{Z \sim \pi} f(g_{\hat{\theta}_{m,n}}(Z))- \hat{\mathbb{E}}^m_{Z \sim \pi} f(g_{\hat{\theta}_{m,n}}(Z))\} + \sup_{||f||_{\text{Lip}}\leq 1} \{ \hat{\mathbb{E}}^m_{Z \sim \pi} f(g_{\hat{\theta}_{m,n}}(Z)) -  \hat{\mathbb{E}}^n_{X \sim \nu} f(X)\} \\ 
  & \hspace{1em} + \sup_{||f||_{\text{Lip}}\leq 1} \{ \hat{\mathbb{E}}^n_{X \sim \nu} f(X) -\mathbb{E}_{X \sim \nu}f(X)\} \\
  \leq & \sup_{||f||_{\text{Lip}}\leq 1} \{\mathbb{E}_{Z \sim \pi} f(g_{\hat{\theta}_{m,n}}(Z))- \hat{\mathbb{E}}^m_{Z \sim \pi} f(g_{\hat{\theta}_{m,n}}(Z))\} + \sup_{||f||_{\text{Lip}}\leq 1} \{ \hat{\mathbb{E}}^m_{Z \sim \pi} f(g_{\theta}(Z)) -  \hat{\mathbb{E}}^n_{X \sim \nu} f(X)\} \\ 
  & \hspace{1em} + \sup_{||f||_{\text{Lip}}\leq 1} \{ \hat{\mathbb{E}}^n_{X \sim \nu} f(X) -\mathbb{E}_{X \sim \nu}f(X)\} \hspace{1em} \text{  it holds for any $g_{\theta} \in \mathcal{G}$ due to \cref{minmax}. } \\
  \leq & \sup_{||f||_{\text{Lip}}\leq 1} \{\mathbb{E}_{Z \sim \pi} f(g_{\hat{\theta}_{m,n}}(Z))- \hat{\mathbb{E}}^m_{Z \sim \pi} f(g_{\hat{\theta}_{m,n}}(Z))\} + \sup_{||f||_{\text{Lip}}\leq 1} \{ \hat{\mathbb{E}}^m_{Z \sim \pi} f(g_{\theta}(Z)) -  \mathbb{E}_{Z \sim \pi} f(g_{\theta}(Z))\}\\
  & \hspace{1em} + \sup_{||f||_{\text{Lip}}\leq 1} \{\mathbb{E}_{Z \sim \pi} f(g_{\theta}(Z))-\mathbb{E}_{X \sim \nu} f(X)\} + 2 \sup_{||f||_{\text{Lip}}\leq 1} \{\mathbb{E}_{X \sim \nu} f(X)-\hat{\mathbb{E}}^n_{X \sim \nu} f(X)\}.
\end{align*}
Note that
\begin{align*}
    \mathbf{Wass}_1(\mu_{\hat{\theta}_{m,n}},\hat{\mu}_{\hat{\theta}_{m,n}}^m)& =\sup_{||f||_{\text{Lip}}\leq 1} \{\mathbb{E}_{Z \sim \pi} f(g_{\hat{\theta}_{m,n}}(Z))- \hat{\mathbb{E}}^m_{Z \sim \pi} f(g_{\hat{\theta}_{m,n}}(Z))\}, \\
    \mathbf{Wass}_1(\mu_{\theta},\hat{\mu}_{\theta}^m) & = \sup_{||f||_{\text{Lip}}\leq 1} \{ \hat{\mathbb{E}}^m_{Z \sim \pi} f(g_{\theta}(Z)) -  \mathbb{E}_{Z \sim \pi} f(g_{\theta}(Z))\}, \\
    \mathbf{Wass}_1(\mu_{\theta},\nu) & = \sup_{||f||_{\text{Lip}}\leq 1} \{\mathbb{E}_{Z \sim \pi} f(g_{\theta}(Z))-\mathbb{E}_{X \sim \nu} f(X)\}, \\
    \mathbf{Wass}_1(\nu,\hat{\nu}^n) & = \sup_{||f||_{\text{Lip}}\leq 1} \{\mathbb{E}_{X \sim \nu} f(X)-\hat{\mathbb{E}}^n_{X \sim \nu} f(X)\},
\end{align*}
where $\hat{\mu_{\theta}}^m$ (or $\hat{\nu}^n$) is an empirical distribution of $m$ (or $n$) i.i.d. samples from $\mu_{\theta}$ 
(or $\nu$).
Because the inequality holds for any $g_{\theta} \in \mathcal{G}$, we have
\begin{align*}
&    \mathbf{Wass}_1(\mu_{\hat{\theta}_{m,n}}, \nu) \\
		& \leq  \mathbf{Wass}_1(\mu_{\hat{\theta}_{m,n}},\hat{\mu}_{\hat{\theta}_{m,n}}^m) + \min_{g_{\theta} \in \mathcal{G}} \{\mathbf{Wass}_1(\mu_{\theta},\hat{\mu}_{\theta}^m) + \mathbf{Wass}_1(\mu_{\theta},\nu)\} + 2 \mathbf{Wass}_1(\nu,\hat{\nu}^n)\\
    & \leq \max_{g_{\theta} \in \mathcal{G}} \mathbf{Wass}_1(\mu_{\theta},\hat{\mu}_{\theta}^m) + \min_{g_{\theta} \in \mathcal{G}} \{\mathbf{Wass}_1(\mu_{\theta},\hat{\mu}_{\theta}^m) + \mathbf{Wass}_1(\mu_{\theta},\nu)\} + 2 \mathbf{Wass}_1(\nu,\hat{\nu}^n).
\end{align*}
By taking expectations on both $m$ i.i.d. samples of $Z$ from $\pi$ and $n$ i.i.d. samples of $X$ from $\nu$, we obtain 
\begin{equation}
\label{bound}
    \begin{aligned}
    & \mathbb{E} \mathbf{Wass}_1(\mu_{\hat{\theta}_{m,n}}, \nu)\\ 
    \leq & \mathbb{E}\max_{g_{\theta} \in \mathcal{G}} \mathbf{Wass}_1(\mu_{\theta},\hat{\mu}_{\theta}^m) +  \mathbb{E} \min_{g_{\theta} \in \mathcal{G}} \{\mathbf{Wass}_1(\mu_{\theta},\hat{\mu}_{\theta}^m) + \mathbf{Wass}_1(\mu_{\theta},\nu)\}+2 \mathbb{E} \mathbf{Wass}_1(\nu,\hat{\nu}^n)\\
    \leq & \mathbb{E}\max_{g_{\theta} \in \mathcal{G}} \mathbf{Wass}_1(\mu_{\theta},\hat{\mu}_{\theta}^m) +  \min_{g_{\theta} \in \mathcal{G}} \{ \mathbb{E} \mathbf{W}_1(\mu_{\theta},\hat{\mu}_{\theta}^m) + \mathbf{Wass}_1(\mu_{\theta},\nu)\}+2 \mathbb{E} \mathbf{Wass}_1(\nu,\hat{\nu}^n)\\
    \leq & \mathbb{E} \max_{g_{\theta} \in \mathcal{G}} \mathbf{Wass}_1(\mu_{\theta},\hat{\mu}_{\theta}^m) + \min_{g_{\theta} \in \mathcal{G}}\mathbf{Wass}_1(\mu_{\theta},\nu) + \max_{g_{\theta} \in \mathcal{G}} \{\mathbb{E} \mathbf{Wass}_1(\mu_{\theta},\hat{\mu}_{\theta}^m)\}+ 2 \mathbb{E} \mathbf{Wass}_1(\nu,\hat{\nu}^n).
    \end{aligned}
\end{equation}
Note that the last inequality holds since
$$
\min_{g_{\theta} \in \mathcal{G}} \{ \mathbb{E} \mathbf{Wass}_1(\mu_{\theta},\hat{\mu}_{\theta}^m)+ \mathbf{Wass}_1(\mu_{\theta},\nu)\} \leq \min_{g_{\theta} \in \mathcal{G}}\mathbf{Wass}_1(\mu_{\theta},\nu) + \max_{g_{\theta} \in \mathcal{G}} \{\mathbb{E} \mathbf{Wass}_1(\mu_{\theta},\hat{\mu}_{\theta}^m)\}.
$$
From \cref{convg_W1} and \cref{assump_3Moment}, we have  
\begin{equation}
\label{exp_nu}
\mathbb{E} \mathbf{Wass}_{1}\left(\nu, \hat{\nu}^n\right) \leq C_1 \cdot\left\{\begin{array}{ll}
n^{-1 / 2}, & d=1, \\
n^{-1 / 2} \log n, & d=2, \\
n^{-1 / d}, & d \geq 3,
\end{array}\right.    
\end{equation}
where the constant $C_1$ depends only on the 3-moment of the target distribution $\nu$ (i.e., $M_3(\nu)$), and it is independent of the number of samples and the model architecture. 

For any $g_{\theta} \in \mathcal{G}$, we have
\begin{equation}
\label{exp_mu}
\begin{aligned}
    \mathbf{Wass}_1(\mu_{\theta},\hat{\mu}_{\theta}^m) & = \sup_{||f||_{\text{Lip}}\leq 1} \{\mathbb{E}_{Z \in \pi}f(g_{\theta}(Z)) - \hat{\mathbb{E}}^m_{Z \in \pi}f(g_{\theta}(Z))\} \\
    & \leq \sup_{||f||_{\text{Lip}} \leq L} \{\mathbb{E}_{Z \in \pi}f(Z) - \hat{\mathbb{E}}^m_{Z \in \pi}f(Z)\}\\
    & = L \cdot \mathbf{Wass}_1(\pi,\hat{\pi}^m).
\end{aligned}    
\end{equation}
The inequality holds because for any $g_{\theta} \in \mathcal{G}$, $\|g_{\theta}\|_{\text{Lip}} \leq L$, then $f \circ g_{\theta}$ is also $L$-Lipschitz given that $f$ is 1-Lipschitz. Similarly, we have
\begin{equation}
\label{exp_pi}
\mathbb{E} \mathbf{Wass}_{1}\left(\pi, \hat{\pi}^m\right) \leq C_2 \cdot\left\{\begin{array}{ll}
m^{-1 / 2}, & r=1, \\
m^{-1 / 2} \log m, & r=2, \\
m^{-1 / r}, & r \geq 3,
\end{array}\right.    
\end{equation}
where the constant $C_2$ depends only on the 3-moment of the target distribution $\pi$ (i.e., $M_3(\pi)$), regardless of the number of samples. Combining \cref{bound}, \cref{exp_nu}, \cref{exp_mu} and \cref{exp_pi}, we complete the proof of \cref{orig_thm}. 
\end{proof}

%If their empirical distributions are not convergent, let alone %distributions for obtained generators. That is the main reason %for the difficulties of generalization analysis for vanilla GAN.
%\begin{assumption}
%\label{assump_3Moment}
%Both the $d$ dimensional target distribution $\nu$ and the r %dimensional source distribution $\pi$ have finite 3-moment, %i.e. $M_3(\nu)=\mathbb{E}_{X \sim \nu}\|X\|^{3}<\infty$ and %$M_3(\pi)=\mathbb{E}_{Z \sim \pi}\|Z\|^{3}<\infty$.
%\end{assumption}
 
%The proof of \cref{orig_thm} can be found in  %\cref{proof_thm1}. 

The following lemma can be used to 
estimate for $\min_{g_{\theta} \in \mathcal{G}}\mathbf{Wass}_1(\mu_{\theta},\nu)$. 

\begin{lemma}
[Capacity of deep generative networks to approximate distributions \cite{yang2022capacity}.]
\label{lemma_capacity}
Let $p \in[1, \infty)$ and $\pi$ be an absolutely continuous probability distribution on $\mathbb{R}^{r}$. Assume that the target distribution $\nu$ is on $\mathbb{R}^{d}$ with finite absolute $q$-moment $M_{q}(\nu)<\infty$ for some $q>p .$ Suppose that $W_g \geq 7 d+1$ and $D_g \geq 2 .$ Then there exists a neural network $g_{\theta}(\mathbf{x}) = \text{NN}(\mathbf{x}|W_g,D_g,\theta)$ such that the generated distribution $\mu_{\theta}$ of $g_{\theta}(Z)$ with $Z \sim \pi$ has the following property:
$$
\mathbf{Wass}_{p}\left(\mu_{\theta}, \nu \right) \leq C \cdot \left(M_{q}^{q}(\nu)+1\right)^{1 / p}\left\{\begin{array}{ll}
\left(W_g^{2} D_g\right)^{-1 / d}, & q>p+p / d, \\
\left(W_g^{2} D_g\right)^{-1 / d}\left(\log _{2} W_g^{2} D_g\right)^{1 / d}, & p<q \leq p+p / d,
\end{array}\right.
$$
where $C$ is a constant depending only on $p, q$ and $d$.
\end{lemma}

%\begin{proof}
See \cite{yang2022capacity} for the detailed proof, the idea of which is first to split $\mathbf{Wass}_{p}\left(\mu_{\theta}, \nu \right)$ into two terms, i.e., $\mathbf{Wass}_{p}\left(\mu_{\theta}, \nu \right) \leq \mathbf{Wass}_{p}\left(\mu_{\theta}, \hat{\nu}^n \right)+\mathbf{Wass}_{p}\left(\hat{\nu}^n, \nu \right)$, then construct a special piece-wise linear function that can be realized by the class of ReLU NNs (i.e., generators $g_{\theta}(\mathbf{x}) = \text{NN}(\mathbf{x}|W_g,D_g,\theta)$ defined in \cref{generator_class}) with $\mathbf{Wass}_{p}\left(\mu_{\theta}, \hat{\nu}^n \right)=0$, if $W_g$ and $D_g$ are sufficiently large compared with $n$. Furthermore, $\mathbf{Wass}_{p}\left(\hat{\nu}^n, \nu \right) \to 0$ as $n \to \infty$ using the generalization property of Wasserstein-$p$ distance. 
The proof is completed by leveraging both two terms with $W_g$
and $D_g$.
		%andcompletes the proof.}
%\end{proof}

Although Yang et al. \cite{yang2022capacity} proved \cref{lemma_capacity}  for a one-dimensional source distribution (i.e., $r=1$), the result can be extended to $r > 1$. 
The main reason for this is that the distribution obtained by a linear projection on absolutely continuous distributions is still absolutely continuous.
%high dimensional absolutely continuous source distributions %simply by linear projections. In other words, linear %combinations of absolutely continuous distributions are still %absolutely continuous. 
This is the reason why the upper bound in \cref{lemma_capacity} does not involve the input dimension $r$. 
In general, it is challenging to make a significant improvement to the upper bound with larger $r$, but a
suitable value of $r$ is preferred 
in practice because 
of the following 
%rather than as large as possible is preferred, mainly due to the 
intuitive phenomenon 
%that
\begin{equation*}
        \mathop{\lim \inf}_{r \to \infty} \min_{\theta}\mathbf{Wass}_{p}\left(\mu_{\theta}, \nu \right) \geq {\rm a \ constant} \ > 0,
\end{equation*}
for NNs with limited width and depth (capacity).
Therefore, we do not plan to 
%don't 
improve the upper bound in \cref{lemma_capacity} for $r>1$ but it remains a possible research topic.
%is beyond our scope, where we mainly focus on the %generalization of WGAN.

Note that 
\cref{lemma_capacity} provides an estimation for $\min_{g_{\theta} \in \mathcal{G}}\mathbf{Wass}_1(\mu_{\theta},\nu)$. 
According to \cref{lemma_capacity}, if we set $p=1$ and $q=3$, then there exists a generative net that transforms the source distribution $\pi$ to a distribution that is close to the target distribution $\nu$ in $\textbf{Wass}_1$ distance.  Let $W_g$, $D_g$, $L$, and $M$ be sufficiently large, $\min_{g_{\theta} \in \mathcal{G}}\mathbf{Wass}_1(\mu_{\theta},\nu)$ can be arbitrarily small, i.e., with sufficiently large $W_g$, $D_g$ , $L$, and $M$, we have
$$
\min_{g_{\theta} \in \mathcal{G}}\mathbf{Wass}_1(\mu_{\theta},\nu) \approx 0.
$$
Furthermore, if we fix the generator class $\mathcal{G}$ ($W_g$, $D_g$, $L$ and $M$ are fixed) and let $m$ and $n$ to be sufficiently large, then by \cref{bound_final} we have
$$ 
\mathbb{E} \mathbf{Wass}_1(\mu_{\hat{\theta}_{m,n}}, \nu) \approx \min_{g_{\theta} \in \mathcal{G}}\mathbf{Wass}_1(\mu_{\theta},\nu) \approx 0.$$
Because $\mathbf{Wass}_1(\mu_{\hat{\theta}_{m,n}}, \nu) \geq 0$, we deduce that
$$
\mathbf{Wass}_1(\mu_{\hat{\theta}_{m,n}}, \nu) \approx 0, \text{ for most of the cases (with a high probability).}
$$

% however, for a concrete bound of $\mathbb{E} \mathbf{Wass}_1(\mu_{\hat{\theta}_{m,n}}, \nu)$, we need further assumptions because the generator class in \cref{lemma_capacity} is not bounded (i.e., $L$ and $M$ are infinite). For a fixed target distribution $\nu$ and the source distribution $\pi$, it's reasonable to assume that \cref{lemma_capacity} holds for the generator class with bounded Lipschitzness and bias. In other words, there exist large enough $L$ and $M$ such that \cref{lemma_capacity} holds for generators defined in \cref{generator_class} for all large enough $D_g$ and $W_g$. More details are illustrated in the \cref{assump_bound_gen}.

\begin{assumption}
\label{assump_bound_gen}
% Suppose that for the fixed target distribution $\nu$ and the source distribution $\pi$, there exist large enough Lipschitz constant $L>0$ and bias $M>0$, such that \cref{lemma_capacity} holds for the generator class defined in \cref{generator_class}.
Suppose that for the fixed target distribution $\nu$ and the source distribution $\pi$, the neural network $g_{\theta}(\mathbf{x})$ satisfying \cref{lemma_capacity} is $L$-Lipschitz and has bounded bias, i.e., there exists $M>0$ such that $\|g_{\theta}(\mathbf{0})\|_2 \leq M$. 
\end{assumption}

\cref{assump_bound_gen} of Lipschitzness and bounded bias is common and essential in generalization analysis; e.g., 
see  \cite{arora2017generalization,bai2018approximability,liang2021well}. For target distributions with a large mean (bias), we usually first preprocess the data by zero mean or normalization. Therefore, the requirement of bounded bias for generators does not influence their approximation capabilities. Furthermore, the assumption of bounded Lipschitzness for NNs is common and natural in convergence and generalization analysis \cite{arora2017generalization, bai2018approximability}. If the optimal transport from the source distribution to the target distribution in $\mathbf{Wass}_1$ is Lipschitz continuous with high probabilities, then we expect NNs (generators) with bounded Lipschitzness ($L$ is sufficiently large) to approximate the implicit optimal transport well. Also, bounded Lipschitzness contributes to the stable training of deep learning models, not only for GANs \cite{pauli2021training, erichson2021lipschitz}. The above analysis accounts for the reasonableness of \cref{assump_bound_gen}. Combining with \cref{assump_bound_gen}, \cref{orig_thm} is rewritten as the following \cref{thm_nonpara_bound}.

\begin{theorem}
[Generalization error bound for WGAN with nonparametric discriminators
in terms of width and depth of generators.]
\label{thm_nonpara_bound}
Suppose that \cref{assump_3Moment} holds for  the target distribution $\nu$ (lies on $\mathbb{R}^d$) and the source distribution $\pi$ (lies on $\mathbb{R}^r$). For GANs, the generator class $\mathcal{G}$ is defined in \cref{generator_class} and the discriminator class $\mathcal{F}=\mathcal{F}_W$ (defined in \cref{W_discriminator_class}), i.e., WGANs. Furthermore, assume that \cref{assump_bound_gen} holds for generators defined in \cref{generator_class} with sufficiently large $L>0$ and $M>0$. If the set of parameters $\hat{\theta}_{m,n}$ is obtained from \cref{minmax}, then the following holds,
\begin{equation}
\label{nonp_error_bound}
\begin{split}
& \mathbb{E} \mathbf{Wass}_1(\mu_{\hat{\theta}_{m,n}}, \nu)\\
\leq & C \cdot W_g^{-2/d}D_g^{-1/d}+C_1\cdot L \cdot \left\{\begin{array}{ll}
m^{-1 / 2}, & r=1, \\
m^{-1 / 2} \log m, & r=2,\\
m^{-1 / r}, & r \geq 3,
\end{array}\right. + C_2\cdot\left\{\begin{array}{ll}
n^{-1 / 2}, & d=1, \\
n^{-1 / 2} \log n, & d=2, \\
n^{-1 / d}, & d \geq 3,
\end{array}\right.    
\end{split}
\end{equation}
for sufficiently large $m$ and $n$, where the constant $C$ depends on $M_3(\nu)$ and $d$, and the constants $C_1$ and $C_2$ depends only on $M_3(\pi)$ and $M_3(\nu)$, respectively.
\end{theorem}

\begin{proof}
From \cref{orig_thm} and \cref{assump_bound_gen}, the inequality \cref{nonp_error_bound} is easily obtained. In summary, \cref{thm_nonpara_bound} implies that with sufficient many training data ($m$ and $n$ are sufficiently large), the obtained generated distribution $\mu_{\hat{\theta}_{m,n}}$ is close to the target distribution $\nu$ in $\mathbf{Wass}_1$ distance, if the generator class $\mathcal{G}$ defined in \cref{generator_class} is of large capacity ($W_g$ and $D_g$ are sufficiently large).
\end{proof}

\subsection{Parametric discriminators (GroupSort neural networks)}
\label{generalization_para}

In practice, we cannot use $\mathcal{F}_W$ as the discriminator class of WGANs, because it would be difficult to handle the whole class of 1-Lipschitz functions. 
As noted in \cref{w_distance_wgan}, both theoretical and numerical results (see in \cite{anil2019sorting, tanielian2021approximating}) show that GroupSort NNs are stronger than ReLU or LeakyReLU NNs in approximating 1-Lipschitz functions. Here, we use GroupSort NNs as the discriminator class, i.e., $\mathcal{F}=\mathcal{F}_{\text{GS}}$ (defined in \cref{discriminator_class}). 

In this subsection, we take the discriminator class $\mathcal{F}_{\text{GS}}$  as a surrogate for $\mathcal{F}_W$ (defined in \cref{W_discriminator_class}) and obtain the estimated generator $g_{\hat{\theta}_{m,n}}$ from \cref{minmax}. Note that condition \cref{constraints} guarantees that all  GroupSort NNs in $\mathcal{F}_{\text{GS}}$ are 1-Lipschitz because the GroupSort activation function is norm-preserving (see \cref{1lipschitzness}).
Next, we require (polynomial) decaying conditions for both the generated distribution and the target distributions; more precisely, their tails of the generated distribution and the target distribution are at least polynomially decaying. Note that when mapped by $L$-Lipschitz functions (generators), the tail of the generated distribution retains the shape of the source distribution \cite{huster2021pareto}. 

\begin{assumption}[Decaying condition]
\label{decay}
Suppose that for a large enough number $\widetilde{M}$ and $s \geq 2$, there exists a constant $C$  such that 
$$
%\begin{align*}
    \mathbb{P}\left ( || g_{\theta}(Z)||_2 \geq \widetilde{M}\right ) 
		\leq C \cdot L^s \widetilde{M}^{-s}
		%\\
\quad {\rm and} \quad    \mathbb{P}\left ( \left \lVert X\right \rVert_2 \geq \widetilde{M}\right ) 
%& 
\leq C \cdot \widetilde{M}^{-s}
$$
%\end{align*}
hold for $X \sim \nu$ and all $g_{\theta} \in \mathcal{G}$ with $Z \sim \pi$. Here, we require $\widetilde{M} \geq 2M$ and $\widetilde{M} \geq L$, and the constant $C$ depends only on the two distributions $\nu$ and $\pi$.
\end{assumption}

The decaying conditions for the target distribution and the generated distributions are mild. In practice, we can verify that the data bias $M$ is usually very close to zero after data normalization (i.e., data preprocessing, which is essential in deep learning). In image generation, image data are bounded and are usually scaled to $[-1,1]$, and even heavy-tailed social-network data  (e.g., Wikipedia Web traffic and SNAP LiveJournal \cite{leskovec2009community}) still follow polynomial decay. As for the generated distribution, if the source distribution $Z \sim \pi$ is standard Gaussian $\mathcal{N}(\mathbf{0},\mathbf{I}_{r})$ (one of the most accepted source distributions in GANs), then for all $g_{\theta} \in \mathcal{G}$ (which is $L$-Lipschitz and with bias $M$), $g_{\theta}(Z)$ is sub-Gaussian. This further validates the conclusion that $L$-Lipschitz mappings do not change the shape of distribution tails. More concretely, for any $g_{\theta} \in \mathcal{G}$, we have
$$
||g_{\theta}(Z)||_2 \leq L||Z||_2+ M,
$$
therefore, 
\begin{equation}
\label{decay_gaussian}
    \mathbb{P}\left ( || g_{\theta}(Z)||_2 \geq \widetilde{M}\right ) \leq \mathbb{P}\left ( || Z||_2 \geq L^{-1}(\widetilde{M}-M)\right ) \leq  \mathbb{P}\left ( || Z||_2 \geq \frac{1}{2} L^{-1}\widetilde{M}\right ) 
\end{equation}
when $\widetilde{M} \geq 2M $. The inequality \cref{decay_gaussian} implies that $g_{\theta}(Z)$ and $Z$ share the same tail shapes. If $Z$ is from a standard Gaussian distribution, then there exists a universal constant $C$ dependent on $L$ and $M$, such that
$$
\mathbb{P}\left ( || Z||_2 \geq \frac{1}{2}L^{-1}\widetilde{M}\right )  \leq C \cdot L^{s} \widetilde{M}^{-s},
$$
because $\mathbb{P}\left ( || Z||_2 \geq \widetilde{M}\right )  \lesssim  exp(-\widetilde{M}^2/2r)$.
Any bounded distribution (i.e., the random variables are bounded almost everywhere) apparently satisfies the above assumption. Moreover, if the decaying condition holds with $s>3$, then \cref{assump_3Moment} for the finite 3-moment of the target distribution $\nu$ and the source distribution $\pi$ is automatically satisfied.

\begin{theorem}
[Generalization error bound for WGANs with GroupSort networks as discriminators.]
% in approximating distributions by $1$-Lipschitzness of %GroupSort neural networks]
\label{main_thm}
Suppose that \cref{assump_3Moment} holds for the target distribution $\nu$ (lies on $\mathbb{R}^d$) and the source distribution $\pi$ (lies on $\mathbb{R}^r$). For GANs, the generator class $\mathcal{G}$ is defined in \cref{generator_class}, and the discriminators are GroupSort NNs (i.e., $\mathcal{F}=\mathcal{F}_{\text{GS}}$) in the form of \cref{groupsortnn} with constraints \cref{constraints}, i.e., GroupSort WGANs. Furthermore, assume that \cref{assump_bound_gen} holds for the generator class $\mathcal{G}$ and \cref{decay} holds for $s = 2$ and $\widetilde{M} = L \cdot (2^{D_f/(2d^2)} \wedge  W_f^{1/(2d^2)})$. If the set of parameters $\hat{\theta}_{m,n}$ is obtained from \cref{minmax}, then we have the following generalization error bound  for the generated distribution $\mu_{\hat{\theta}_{m,n}}$,
\begin{equation}
\label{generalization_thm}
\begin{split}
    \mathbb{E}\mathbf{Wass}_1(\mu_{\hat{\theta}_{m,n}},\nu) \lesssim \hspace{0.5em} &  W_g^{-2/d}D_g^{-1/d}+ L \cdot (m^{-1/2}\log m \vee m^{-1/r})+(n^{-1/2} \log n \vee n^{-1/d}) \\
& +  L \cdot ( 2^{-D_f/(2d^2)} \vee W_f^{-1/(2d^2)}),
\end{split}
\end{equation}
when $m$, $n$, $D_f$ and $W_f$ are sufficiently large. 
Here, $\lesssim$ hides some parameters related to 
%$d$, $r$, 
$M_3(\nu)$ and $M_3(\pi)$.
% but independent of the model architecture. 
\end{theorem}

Sketch of proof: The generalization error bound is separated into four parts, i.e., (i) the approximation error of deep generative networks, (ii) the convergence of generated samples, (iii) the convergence of samples from target distributions, and (iv) the approximation capabilities of GroupSort NNs to 1-Lipschitz functions. The detailed proof is given later;
%\cref{proof_thm3}. 
here we briefly explain some intuitions from \cref{generalization_thm} here.
\begin{itemize}
    \item The first term in the right-hand side of \cref{generalization_thm} represents the expressive power of generators. If we constrain the Lipschitzness and bias of generators ($L$ and $M$ are sufficiently large), then the capacity of generators improves with increasing width and depth (with \cref{assump_bound_gen}).
    \item The second term is the generalization error of generators with only a finite number of generated samples. As can be seen, this term is actually related to the diversity of generators, which is the Lipschitzness we constrain.
    \item The third term denotes the generalization error from the target distribution due to the limited number of observed data.
    \item The last term comes from the approximation of GroupSort NNs to 1-Lipschitz functions. Here, we require that \cref{decay} to hold for $s = 2$ and $\widetilde{M} = L \cdot (2^{D_f/(2d^2)} \wedge  W_f^{1/(2d^2)})$, which is mild. Furthermore, we require that $L \ll (2^{D_f/(2d^2)} \wedge  W_f^{1/(2d^2)})$, otherwise, we observe a large error that is dominated by $L$.
\end{itemize}
The generalization error bound is small if the width and depth of discriminators are sufficiently large (the error decays exponentially and polynomially with depth and width, respectively), which implies that the capacities of discriminators contribute to generation quality.
% Noted that $M^{2D_g} \lesssim \widetilde{M} = W_f^{1/2d^2} \wedge D_f$ shows severe curse of dimensionality in Wasserstein GAN's error bound and it also means that GANs require strong discriminators, consistent with some empirical and theoretical results.
% Moreover, $W_g  \leq M^{1/2}$ and $M^{2D_g} \lesssim \widetilde{M} = W_f^{1/2d^2} \wedge D_f$ also indicate strict requirement for discriminators in GANs. It implies that overly deep and wide (high capacity) generators may cause worse results than low capacity generators if discriminators are not strong enough.
% Based on this, we have reasons to believe that the mode collapse in GANs may attribute to low capacity of discriminator class. The main difference between Theorem \ref{thm_nonpara_bound} and Theorem \ref{main_thm} is the error from the approximation of $\mathcal{F}_{GS}$ to 1-Lipschitz functions.

We first consider the following proposition before we give the proof
of \cref{main_thm}.

\begin{proposition}[Lemma 1, \cite{tanielian2021approximating}]
\label{1lipschitzness}
Assume that \cref{constraints}  is satisfied for GroupSort neural networks. Then, for any $k \geqslant 2$ ($k$ is the grouping size for GroupSort activation functions in \cref{groupsortnn}), $\mathcal{F}_{\text{GS}}$ defined in \cref{discriminator_class} is a subset of 1-Lipschitz functions defined on $\mathbb{R}^d$, i.e., $\mathcal{F}_{\text{GS}} \subseteq \mathcal{F}_W$.
\end{proposition}

For any $f \in \mathcal{F}_W = \{f: ||f||_{\text{Lip}}\leq 1\}$ and supposing that $g_{\hat{\theta}_{m,n}}$ is obtained from \cref{minmax}, we have 
\begin{align*}
    &  \hspace{1em} \mathbb{E}_{Z \sim \pi} f(g_{\hat{\theta}_{m,n}}(Z))-\mathbb{E}_{X \sim \nu}f(X) \\
    & = \mathbb{E}_{Z \sim \pi} f(g_{\hat{\theta}_{m,n}}(Z))- \mathbb{E}_{Z \sim \pi} f_{\alpha}(g_{\hat{\theta}_{m,n}}(Z))+\mathbb{E}_{Z \sim \pi} f_{\alpha}(g_{\hat{\theta}_{m,n}}(Z)) - \hat{\mathbb{E}}^m_{Z \sim \pi} f_{\alpha}(g_{\hat{\theta}_{m,n}}(Z))\\
    & + \hat{\mathbb{E}}^m_{Z \sim \pi} f_{\alpha}(g_{\hat{\theta}_{m,n}}(Z)) - \hat{\mathbb{E}}_{X \sim \nu}^n f_{\alpha}(X)+ \hat{\mathbb{E}}_{X \sim \nu}^n f_{\alpha}(X) - \mathbb{E}_{X \sim \nu} f_{\alpha}(X)+ \mathbb{E}_{X \sim \nu} f_{\alpha}(X)-\mathbb{E}_{X \sim \nu} f(X)\\
    & \leq \min_{f_{\alpha} \in \mathcal{F}} \left \{ \mathbb{E}_{Z \sim \pi} f(g_{\hat{\theta}_{m,n}}(Z))- \mathbb{E}_{Z \sim \pi} f_{\alpha}(g_{\hat{\theta}_{m,n}}(Z)) + \mathbb{E}_{X \sim \nu} f_{\alpha}(X)-\mathbb{E}_{X \sim \nu} f(X) \right\} \\
    & + \max_{f_{\alpha} \in \mathcal{F}} \left \{ \mathbb{E}_{Z \sim \pi} f_{\alpha}(g_{\hat{\theta}_{m,n}}(Z)) - \hat{\mathbb{E}}^m_{Z \sim \pi} f_{\alpha}(g_{\hat{\theta}_{m,n}}(Z)) + \hat{\mathbb{E}}_{X \sim \nu}^n f_{\alpha}(X) - \mathbb{E}_{X \sim \nu} f_{\alpha}(X) \right \}\\
    & + \max_{f_{\alpha} \in \mathcal{F}} \left \{ \hat{\mathbb{E}}^m_{Z \sim \pi} f_{\alpha}(g_{\hat{\theta}_{m,n}}(Z)) - \hat{\mathbb{E}}_{X \sim \nu}^n f_{\alpha}(X) \right \}\\
    & \leq \max_{f_{\alpha} \in \mathcal{F}} \left \{ \mathbb{E}_{Z \sim \pi} f_{\alpha}(g_{\hat{\theta}_{m,n}}(Z)) - \hat{\mathbb{E}}^m_{Z \sim \pi} f_{\alpha}(g_{\hat{\theta}_{m,n}}(Z))\right \} + \max_{f_{\alpha} \in \mathcal{F}} \left \{ \hat{\mathbb{E}}_{X \sim \nu}^n f_{\alpha}(X) - \mathbb{E}_{X \sim \nu} f_{\alpha}(X) \right \}\\
    & + \max_{f_{\alpha} \in \mathcal{F}} \left \{ \hat{\mathbb{E}}^m_{Z \sim \pi} f_{\alpha}(g_{\hat{\theta}_{m,n}}(Z)) - \hat{\mathbb{E}}_{X \sim \nu}^n f_{\alpha}(X) \right \}\\
    & + \min_{f_{\alpha} \in \mathcal{F}} \left \{ \mathbb{E}_{Z \sim \pi} f(g_{\hat{\theta}_{m,n}}(Z))- \mathbb{E}_{Z \sim \pi} f_{\alpha}(g_{\hat{\theta}_{m,n}}(Z)) + \mathbb{E}_{X \sim \nu} f_{\alpha}(X)-\mathbb{E}_{X \sim \nu} f(X) \right\}\\
    & =: I_1 + I_2 + I_3 + I_4,
\end{align*}
where 
\begin{align*}
    I_1 & \leq  \mathbf{Wass}_1(\mu_{\hat{\theta}_{m,n}},\hat{\mu}_{\hat{\theta}_{m,n}}^m)\leq \max_{g_{\theta} \in \mathcal{G}} \mathbf{Wass}_1(\mu_{\theta},\hat{\mu}_{\theta}^m),\\
    I_2 & \leq \mathbf{Wass}_1(\nu,\hat{\nu}^n),\\
    I_3 & = \max_{f_{\alpha} \in \mathcal{F}} \{\hat{\mathbb{E}}^m_{Z \sim \pi} f_{\alpha}(g_{\hat{\theta}_{m,n}}(Z)) - \hat{\mathbb{E}}^n_{X \sim \nu} f_{\alpha}(X)\} \\
    & \leq \max_{f_{\alpha} \in \mathcal{F}} \{\hat{\mathbb{E}}^m_{Z \sim \pi} f_{\alpha}(g_{\theta}(Z)) - \hat{\mathbb{E}}^n_{X \sim \nu} f_{\alpha}(X)\} \hspace{1em} \text{ it holds for all } g_{\theta} \in \mathcal{G} \\
    & \leq \max_{f_{\alpha} \in \mathcal{F}} \{\hat{\mathbb{E}}^m_{Z \sim \pi} f_{\alpha}(g_{\theta}(Z)) -\mathbb{E}_{Z \sim \pi} f_{\alpha}(g_{\theta}(Z))+\mathbb{E}_{Z \sim \pi} f_{\alpha}(g_{\theta}(Z)) - \mathbb{E}_{X \sim \nu} f_{\alpha}(X)\\
    & \hspace{1em} +  \mathbb{E}_{X \sim \nu} f_{\alpha}(X) - \hat{\mathbb{E}}^n_{X \sim \nu} f_{\alpha}(X)\}, \\
   &  \leq \min_{g_{\theta} \in \mathcal{G}} \sup_{||f||_{\text{Lip}} \leq 1} \{\hat{\mathbb{E}}^m_{Z \sim \pi} f(g_{\theta}(Z)) -\mathbb{E}_{Z \sim \pi} f(g_{\theta}(Z))+\mathbb{E}_{Z \sim \pi} f(g_{\theta}(Z)) - \mathbb{E}_{X \sim \nu} f(X) \} \\
	& + \mathbf{Wass}_1(\nu,\hat{\nu}^n) \\
    & \leq \min_{g_{\theta} \in \mathcal{G}} \mathbf{Wass}_1(\mu_{\theta}, \nu) + \max_{g_{\theta} \in \mathcal{G}} \mathbf{Wass}_1(\mu_{\theta},\hat{\mu}_{\theta}^{m}) + \mathbf{Wass}_1(\nu,\hat{\nu}^n), \\
    I_4 & = \min_{f_{\alpha} \in \mathcal{F}} \left \{ \mathbb{E}_{Z \sim \pi} f(g_{\hat{\theta}_{m,n}}(Z))- \mathbb{E}_{Z \sim \pi} f_{\alpha}(g_{\hat{\theta}_{m,n}}(Z)) + \mathbb{E}_{X \sim \nu} f_{\alpha}(X)-\mathbb{E}_{X \sim \nu} f(X) \right\}\\
    & \leq \min_{f_{\alpha} \in \mathcal{F}}\left \{ ||f-f_{\alpha}||_{L^{\infty}([-\widetilde{M},\widetilde{M}]^d)} \cdot \left ( \mathbb{P}\left ( || g_{\hat{\theta}_{m,n}}(Z)||_2 \leq \widetilde{M}\right )+\mathbb{P}\left ( \left \lVert X\right \rVert_2 \leq \widetilde{M}\right ) \right )\right.\\
    & \left. \hspace{1em}+ \mathbb{E}_{Z \sim \pi} [(f(g_{\hat{\theta}_{m,n}}(Z))-f_{\alpha}(g_{\hat{\theta}_{m,n}}(Z)) ) \cdot \mathbb{I}( ||g_{\hat{\theta}_{m,n}}(Z) ||_2 \geq \widetilde{M})] \right. \\
    & \hspace{1em} \left. +\mathbb{E}_{X \sim \nu} [f_{\alpha}(X)-f(X)]\cdot \mathbb{I}(||X||_2 \leq \widetilde{M})\right \}.
\end{align*}
Here, $\mathbb{I}(\cdot)$ is an indicator function.
For a fixed $f \in \{f: ||f||_{\text{Lip}} \leq 1\}$, suppose that $\min_{f_{\alpha} \in \mathcal{F}} ||f-f_{\alpha}||_{L^{\infty}([-\widetilde{M},\widetilde{M}])}  \leq 1$. If $f(0)=0$ and is a $1$-Lipschitz function, then we have $|f_{\alpha}(0)| \leq 1$ and
\begin{align*}
    ||f(X)||_{\infty} & \leq ||X||_2, \\
    ||f_{\alpha}(X)||_{\infty} & \leq ||X||_2+|f_{\alpha}(0)| \leq ||X||_2+1.
\end{align*}
Based on \cref{decay}, we have
\begin{equation}
    \begin{split}
        \mathbb{E}_{Z \sim \pi} [(f(g_{\hat{\theta}_{m,n}}(Z))-f_{\alpha}(g_{\hat{\theta}_{m,n}}(Z)) ) \cdot \mathbb{I}( ||g_{\hat{\theta}_{m,n}}(Z) ||_2 \geq \widetilde{M})]\\
        +\mathbb{E}_{X \sim \nu} [f_{\alpha}(X)-f(X)]\cdot \mathbb{I}(||X||_2 \geq \widetilde{M})] \lesssim L^s\widetilde{M}^{-s+1},
    \end{split}
\end{equation}
where $\lesssim$ hides unimportant constant parameters. The next Lemma mainly discusses the approximation capabilities of GroupSort NNs to 1-Lipschitz functions. The GroupSort NN can approximate any 1-Lipschitz functions arbitrarily well on a compact set if it is sufficiently wide and deep.

\begin{lemma}[Theorem 2, \cite{tanielian2021approximating}]
Let $d \geqslant2$, and the grouping size 
be $k=2$. For any 1-Lipschitz function $f$  defined on $[-\widetilde{M},\widetilde{M}]^{d}$,
there exists a GroupSort neural network $f_{\alpha}$ of form \cref{groupsortnn} with depth $D_f$ and width $W_f$ such that
\begin{equation}
    \|f-f_{\alpha}\|_{L^{\infty}([-\widetilde{M},\widetilde{M}]^{d}))} \lesssim \widetilde{M} \cdot \sqrt{d} \cdot ( 2^{-D_f/(d^2)} \vee W_f^{-1/d^2}).
\end{equation}
\end{lemma}

\begin{proof}
For any 1-Lipschitz function $f$ defined on $[-\widetilde{M},\widetilde{M}]^d$, we can find a 1-Lipschitz function $h$ defined on $[0,1]^d$ satisfying
$$
f(x)=2\widetilde{M}\cdot h(\frac{x}{2\widetilde{M}}+\frac{1}{2}).
$$
By Theorem 2 in \cite{tanielian2021approximating}, there exists a GroupSort NN $h_{\alpha}$ of form \cref{groupsortnn} such that 
$$
||h_{\alpha}-h||_{L^{\infty}([0,1]^d)} \leq \sqrt{d}\cdot (2^{-D_f/(d^2)} \vee W_f^{-1/d^2}).
$$
Let
$$
f_{\alpha}=2\widetilde{M}\cdot h_{\alpha}(\frac{x}{2\widetilde{M}}+\frac{1}{2}) \in \mathcal{F}_{GS}.
$$
Therefore,
$$
\|f-f_{\alpha}\|_{L^{\infty}([-\widetilde{M},\widetilde{M}]^{d}))} \leq 2\widetilde{M} \cdot \|h-\Tilde{h}\|_{L^{\infty}([0,1]^{d}))} \lesssim \widetilde{M} \cdot \sqrt{d} \cdot ( 2^{-D_f/d^2} \vee W_f^{-1/d^2}). 
%\quad \quad \quad \mbox
$$
The lemma follows.
\end{proof}

%Since the probability is less than or equal to 1, 
For the above $f$, we have
\begin{equation}
    I_4 \lesssim \widetilde{M} \cdot  (2^{-D_f/(C d^2)} \vee  W_f^{-1/d^2}) +  L^s \widetilde{M}^{-(s-1)}.
\end{equation}
Here, we omit unimportant parameters related to $d$. Letting $\widetilde{M} = L \cdot (2^{D_f/(s d^2)} \wedge  W_f^{1/sd^2})$ and $s = 2$, we have
\begin{equation}
\label{I4}
    I_4 \lesssim  L \cdot (2^{-D_f/(2 d^2)} \vee W_f^{-1/(2d^2)}).
\end{equation}
Then,
\begin{equation}
\label{bound_thm3}
\begin{split}
    & \hspace{2em} \mathbf{Wass}_1(\mu_{\hat{\theta}_{m,n}},\nu)\\
    & = \sup_{||f||_{\text{Lip}} \leq 1} \{ \mathbb{E}_{Z \sim \pi} f(g_{\hat{\theta}_{m,n}}(Z))-\mathbb{E}_{X \sim \nu}f(X) \}  \leq I_1 + I_2 +I_3 +I_4 \\
    & \leq \min_{g_{\theta} \in \mathcal{G}} \mathbf{Wass}_1(\mu_{\theta}, \nu) + 2 \max_{g_{\theta} \in \mathcal{G}} \mathbf{Wass}_1(\mu_{\theta},\hat{\mu}_{\theta}^m) + 2 \mathbf{Wass}_1(\nu,\hat{\nu}^n) + I_4,
    \end{split}
\end{equation}
To use of the bound \cref{I4} for $I_4$, we need to prove that there exists a 1-Lipschitz function $f$ with $f(0)=0$  that maximizes the term $\mathbb{E}_{Z \sim \pi} f(g_{\hat{\theta}_{m,n}}(Z))-\mathbb{E}_{X \sim \nu}f(X)$. Suppose that $f^{'}$ is one of the maximizers, then $f^{'}-f^{'}(0)$ also maximizes $\mathbb{E}_{Z \sim \pi} f(g_{\hat{\theta}_{m,n}}(Z))-\mathbb{E}_{X \sim \nu}f(X)$ and this finishes the task. 

By taking expectations on both sides of \cref{bound_thm3} and combining it with \cref{bound}, \cref{exp_nu}, \cref{exp_mu}, \cref{exp_pi} and \cref{I4}, we obtain \cref{generalization_thm}
for the proof of \cref{main_thm}.

From the results in \cref{main_thm}, we make the following 
remarks.

\begin{remark}
The above analysis implies that 
$$
\sup_{||f||_{\text{Lip}} \leq 1} 
 \left \{\mathbb{E}_{Z \sim \pi} f(g_{\theta}(Z))-\mathbb{E}_{X \sim \nu}f(X) \right\} = \sup_{||f||_{\text{Lip}} \leq 1, f(0)=0} \left\{ \mathbb{E}_{Z \sim \pi} f(g_{\theta}(Z))-\mathbb{E}_{X \sim \nu}f(X) \right\}.
$$
If $f$ is 1-Lipschitz and $f(0)=0$, then $||f(X)||_2 \leq \sqrt{d} \cdot \widetilde{M}$ on $[-\widetilde{M},\widetilde{M}]^d$, and the bounded function class defined on the compact domain $[-\widetilde{M},\widetilde{M}]^d$ is easier to be approximated. 
\end{remark}

\begin{remark}[Regularization on the Lipschitzness of generators]
\label{remark1}
    There are several ways to constrain the Lipschitzness of NNs in deep learning applications, one of the most common and popular being $L_2$ regularization. Mathematically, there exists $\Gamma>0$ such that $\sum_{i=1}^{D_g}\|\mathbf{W}_i\|_F^2 \leq \Gamma$. Therefore, we have $\sum_{i=1}^{D_g}\|\mathbf{W}_i\|_2^2 \leq \Gamma$ and  $\prod_{i=1}^{D_g}\|\mathbf{W}_i\|_2 \leq \sqrt{\left (\frac{\Gamma}{D_g}\right )^{D_g}}$ by Cauchy–Schwarz inequality. Here, the upper bound for the Lipschitzness of generators satisfies $L \leq \prod_{i=1}^{D_g}\|\mathbf{W}_i\|_2 \leq \sqrt{\left (\frac{\Gamma}{D_g}\right )^{D_g}} \to 0$ when $D_g \to \infty$, which implies that we allow larger $\Gamma$ when NNs are deeper. Similarly, to stabilize GANs training, Brock et al. \cite{brock2016neural} suggested using orthonormal regularization $\sum_{i=1}^{D_g} (\|\mathbf{W}_i^T\mathbf{W}_i - \mathbf{I}\|_F^2$ for generators to softly constrain the Lipschitzness of NNs. The selection of hyperparameters for $L_2$ regularization and orthonormal regularization is tricky and depends on the problem itself. 
    However, \cref{assump_bound_gen} may not hold if with strict regularization on generators. Therefore, we must not only control the Lipschitzness of NNs for stable training but also maintain their expressive power.
\end{remark}

\begin{remark}
    [A case study for generators without regularization]
    \label{remark2}
Suppose that without any regularization, the generator class $\mathcal{G}$ in \cref{generator_class} is represented 
 approximately as 
\begin{equation}
    \label{generator_class2}
    \mathcal{G}=\{g_{\theta}(\mathbf{x})= \text{NN}(\mathbf{x}|W_g,D_g,\theta): \|\mathbf{W}_i\|_{\text{max}}\leq \Gamma \text{ and } \|\mathbf{b}_i\|_{\text{max}}\leq \Gamma,  \theta=\{\mathbf{W}_i,\mathbf{b}_i\}_{i=1}^{D_g} \},
\end{equation}
because the maximal values (i.e., $\|\cdot\|_{\text{max}}$) of the weights and bias are small for NNs by initialization strategies \cite{glorot2010understanding} and tiny learning rates \cite{arjovsky2017wasserstein, anil2019sorting}. Then the Lipschitzness of the generator satisfies $L = (W_g \cdot \Gamma)^{D_g}$ being polynomially and exponentially increasing w.r.t. the width $W_g$ and the depth $D_g$, respectively. Hence, the generalization error bound \cref{generalization_thm_bounded1} further implies that $L = (W_g \cdot \Gamma)^{D_g} \ll  m^{1/r}$ and $L = (W_g \cdot \Gamma)^{D_g} \ll  ( 2^{D_f/(2d^2)} \wedge W_f^{1/(2d^2)})$. Without regularizations on the parameters of generators, their capacity grows with depth and width. Although wider and deeper generators lower the approximation error $W_g^{-2/f}D_g^{-1/d}$, the second and the last terms dominate the generalization error bound due to the limited numbers of training data and weak discriminators. In this sense, the last term further shows that the discriminators should be stronger than generators in WGANs, consistent with some existing numerical observations \cite{arora2017generalization, arjovsky2017wasserstein}. In the experimental part, we report some numerical experiments conducted to test the behaviors of the width and depth of generators and discriminators in GroupSort WGANs. Experimental results further validate the derived generalization error bound \cref{generalization_thm_bounded1} and our analysis here.
\end{remark}

Using \cref{main_thm}, we consider the following special case.

\begin{corollary}
[Generalization error bound for WGANs with GroupSort networks as discriminators under 
bounded target and generated distributions.]
\label{corollary_bounded1}
Suppose that both the target distribution and the generated distribution are bounded by $B>0$. The generator class is defined as 
\begin{equation*}
\mathcal{G} = \{g_{\theta}(\mathbf{x})= \text{NN}(\mathbf{x}|W_g,D_g,\theta): \|g_{\theta}(\mathbf{x})\|_2 \leq B,  \theta=\{\mathbf{W}_i,\mathbf{b}_i\}_{i=1}^{D_g} \},
\end{equation*}
while the discriminators are GroupSort NNs (i.e., $\mathcal{F}=\mathcal{F}_{\text{GS}}$) in the form of \cref{groupsortnn} with constraints \cref{constraints}. Furthermore, assume that \cref{assump_bound_gen} holds for the generator class $\mathcal{G}$. If the set of parameters $\hat{\theta}_{m,n}$ is obtained from \cref{minmax}, then we have the following generalization error bound  for the generated distribution $\mu_{\hat{\theta}_{m,n}}$,
\begin{equation}
\label{generalization_thm_bounded1}
\begin{split}
    \mathbb{E}\mathbf{Wass}_1(\mu_{\hat{\theta}_{m,n}},\nu) \lesssim \hspace{0.5em} &  W_g^{-2/d}D_g^{-1/d}+ B \cdot (m^{-1/2}\log m \vee m^{-1/r})+ B \cdot (n^{-1/2} \log n \vee n^{-1/d}) \\
& +  B \cdot ( 2^{-D_f/(2d^2)} \vee W_f^{-1/(2d^2)}).
\end{split}
\end{equation}
Here, $\lesssim$ hides unimportant parameters related to $d$ and $r$.
%, but independent of the model architecture.  
\end{corollary}

\begin{proof}
The proof is similar to that of  \cref{main_thm}. 
Letting $\widetilde{M}=B$, then we have  
$$
%\begin{align*}
    \mathbb{P}\left ( || g_{\theta}(Z)||_2 \geq \widetilde{M}\right ) 
		=0
		%\\
\quad {\rm and} \quad    \mathbb{P}\left ( \left \lVert X\right \rVert_2 \geq \widetilde{M}\right ) 
%& 
=0.
$$
Therefore, $I_4$ in \cref{I4} is replaced by
$$
%\begin{equation*}
    I_4 \lesssim  B \cdot (2^{-D_f/(2d^2)} \vee W_f^{-1/(2d^2)}),
%\end{equation*}
$$
and the result follows.
\end{proof}

\subsection{The Generalization error bound by Rademacher complexity} 

In \cref{main_thm}, the generalization error bound 
%is undesirable and not strict in the sense of convergence for the numbers of training %data $m$ and $n$, where it 
suffers from the curse of dimensionality (the bound contains
$\mathcal{O}(m^{-1/r})$ and $\mathcal{O}(n^{-1/d})$). Therefore, it would be interesting to provide 
%We prefer 
a generalization error bound with respect to 
$m$ and $n$ with $\widetilde{\mathcal{O}}(m^{-1/2})$ and $\widetilde{\mathcal{O}}(n^{-1/2})$.
In this subsection,
we consider using Rademacher complexity to analyze the generalization
error bound.

%, in deep learning theory literature, while we achieve $\mathcal{O}(m^{-1/r})$ and %$\mathcal{O}(n^{-1/d})$ which are undesirable. In the next theorem, we derive a %new generalization error bound which is of $\widetilde{\mathcal{O}}(m^{-1/2})$ %and $\widetilde{\mathcal{O}}(n^{-1/2})$, for the obtained generated distribution %$\mu_{\hat{\theta}_{m,n}}$. However, we found an interesting phenomenon, we %will discuss it just followed by the theorem. 

\begin{definition}
\label{assump_dis}
For $n$ i.i.d. variables $X_1$, $\cdots$, $X_n$ from $\nu$, the function $F(\nu,n)$ is defined as the expectation of maximal values over those variables. Mathematically, 
\begin{equation}
    F(\nu, n) = \mathbb{E}_{X_1, \cdots, X_n \sim \nu} \max\{\|X_1\|_2, \cdots, \|X_n\|_2\}.
\end{equation}
For a bounded distribution $\nu$, $F(\nu, n)$ is bounded and independent 
with $n$. If the tail of distribution $\nu$  decays exponentially, then $F(\nu, n)$ is proportional to $\log(n)$. For example, if $\nu=\text{Unif}([0,1]^{r})$, then $F(\nu, n) \leq \sqrt{r}$; for standard Gaussian distribution $\nu=\mathcal{N}(\mathbf{0},\mathbf{I}_r)$, we have $F(\nu, n) \leq \sqrt{2r \cdot \log(2n)}$.
\end{definition}

\begin{theorem}
[Generalization error bound for WGANs with GroupSort networks as discriminators by Rademacher compleixty]
\label{new_generalization_thm}
Suppose that \cref{assump_3Moment} holds for the target distribution $\nu$ (lies on $\mathbb{R}^d$) and the source distribution $\pi$ (lies on $\mathbb{R}^r$). For GANs, the generator class $\mathcal{G}$ is defined in \cref{generator_class}
while discriminators are GroupSort NNs (i.e., $\mathcal{F}=\mathcal{F}_{\text{GS}}$) in the form of \cref{groupsortnn} with constraints \cref{constraints}, i.e., GroupSort WGANs. Furthermore, assume that \cref{assump_bound_gen} holds for the generator class $\mathcal{G}$ and \cref{decay} holds for  $s = 2$ and $\widetilde{M} = L \cdot (2^{D_f/(2 d^2)} \wedge  W_f^{1/(2d^2)})$. If the set of parameters $\hat{\theta}_{m,n}$ is obtained from \cref{minmax}, then we have the following generalization error bound for the generated distribution $\mu_{\hat{\theta}_{m,n}}$,
\begin{equation}
\label{generalization_thm2}
\begin{split}
    \mathbb{E}\mathbf{Wass}_1(\mu_{\hat{\theta}_{m,n}},\nu) \lesssim \hspace{0.5em} &  W_g^{-2/d}D_g^{-1/d} + (L \cdot F(\pi,m)+M) \cdot \sqrt{\text{Pdim}(\mathcal{F}_{\text{GS}} \circ \mathcal{G} ) \cdot  \frac{\log m}{m}} \\ & +  F(\nu,n) \cdot \sqrt{\text{Pdim}(\mathcal{F}_{\text{GS}}) \cdot  \frac{\log n}{n}} + L \cdot ( 2^{-D_f/(2d^2)} \vee W_f^{-1/(2d^2)}).
\end{split}
\end{equation}
Here, $\lesssim$ hides parameters related to 
some unimportant universal constants. 
\end{theorem}

Sketch of proof: The proof is similar to that of \cref{main_thm}.
Again, the generalization error bound is separated into four parts: (i) the approximation error of deep generative networks, (ii) the convergence of generated samples, (iii) the convergence of samples from target distributions, and (iv) the approximation of GroupSort NNs to 1-Lipschitz functions. 
The proof of \cref{new_generalization_thm} mainly involves deriving new error bounds for $I_1$ and $I_2$ due to the finite capacity of $\mathcal{F}_{\text{GS}}$. To understand the details fully, some definitions (e.g., covering balls, covering numbers, Rademacher complexity, pseudo-dimension) and theorems (e.g., Dudley's theorem) are required; see \cite{ma2021lecturenotes} for 
a detailed discussion of these.

We use $\mathcal{R}_{\mathcal{S}}(\mathcal{F})$ and $\mathcal{R}_{n}(\mathcal{F})$ to denote the empirical Rademacher complexity of the function space $\mathcal{F}$ on the i.i.d. data $\mathcal{S}=\{X_1, \cdots, X_n\}$ and the corresponding Rademacher complexity with $\mathcal{R}_{n}(\mathcal{F})=\mathbb{E}_{\mathcal{S}} \mathcal{R}_{\mathcal{S}}(\mathcal{F})$, respectively. Here, the empirical Rademacher complexity is defined as 
$$
\mathcal{R}_{\mathcal{S}}(\mathcal{F}):=\mathbb{E}_{\epsilon} \sup _{f \in \mathcal{F}} \frac{1}{n} \sum_{i=1}^{n} \epsilon_{i} f\left(X_{i}\right),
$$
where $\epsilon=\left(\epsilon_{1}, \ldots, \epsilon_{n}\right)$ are i.i.d. Rademacher random variables. \cref{rademacher_complexity}-\cref{lemma_cov_pseudo} show that the generalization error is bounded by the Rademacher complexity of the function class and then is further controlled by the pseudo-dimension.

\begin{lemma}
[Translating the generalization error to the Rademacher complexity.]
\label{rademacher_complexity}
Let $\mathcal{S}=\left \{X_1, \cdots, X_n\right \}$ be a set of $n$ i.i.d. samples, then we have
$$
\mathbb{E}_{\mathcal{S}} \left [ \sup_{f \in \mathcal{F}} \left [ \frac{1}{n} \sum_{i=1}^{n} f(X_i) - \mathbb{E}_{X} \left [f(X) \right ]  \right ] \right ]\leq 2 \mathcal{R}_{n}(\mathcal{F}).
$$
\end{lemma}
For more details about \cref{rademacher_complexity}, see Theorem 4.13 in \cite{ma2021lecturenotes}. As given in \cref{lemma_dudley}, Dudley's Theorem (Theorem 4.26 in \cite{ma2021lecturenotes}) is useful in deep learning theory in that it connects the covering number to the Rademacher complexity.

\begin{lemma}
[Translating the Rademacher complexity to the covering number, Dudley’s Theorem.]
\label{lemma_dudley}If $\mathcal{F}$ is a function class from $\mathcal{X} \to \mathbb{R}$, where $\mathcal{S}$ is sampled from $\mathcal{X}$, and for $B>0$, we assume that $\max_{i} |f(X_i)| \leq B$ for all $f \in \mathcal{F}$. Then,
\begin{equation}
\label{bound_rademacher_complexity}
    R_{S}(\mathcal{F}) \leq 12 \int_{0}^{B} \sqrt{\frac{\log N\left(\epsilon, \mathcal{F}, \|\cdot\|_{2,n} \right)}{n}} d \epsilon,
\end{equation}
where $\|f\|_n = \sqrt{\frac{1}{n}\sum_{i=1}^{2,n}\left(f(X_i)\right)^2}$ and $N \left(\epsilon, \mathcal{F}, \|\cdot\|_{2,n} \right)$ is the covering number for $\mathcal{F}$ with radius $\epsilon$ and metric $\|\cdot\|_{2,n}$.
\end{lemma}

\begin{lemma}
[Translating the covering number to the pseudo-dimension \cite{anthony1999neural}.]
\label{lemma_cov_pseudo}
For $B>0$, assume that  for all $f \in \mathcal{F}$, we have $\max_{i} |f(X_i)| \leq B$. Define the metric $\|\cdot\|_{\infty,n}$ as $\|f\|_{\infty,n}=\max_{i\in \{1, \cdots, n\}}|f(X_i)|$. Then,
\begin{equation}
\label{bound_cover_number_1}
    N\left(\epsilon,\mathcal{F}, \|\cdot\|_{\infty,n}\right) \leq\left(\frac{2 e B \cdot n}{\epsilon \cdot \operatorname{Pdim}(\mathcal{F})}\right)^{\operatorname{Pdim}(\mathcal{F})}
\end{equation}
for any $\epsilon>0$, where $\operatorname{Pdim}(\mathcal{F})$ denotes the pseudo-dimension of $\mathcal{F}$ and $N \left(\epsilon, \mathcal{F}, \|\cdot\|_{\infty,n} \right)$ is the covering number for $\mathcal{F}$ with radius $\epsilon$ and metric $\|\cdot\|_{\infty,n}$.
\end{lemma}

Obviously $N\left(\epsilon,\mathcal{F}, \|\cdot\|_{2,n}\right) \leq N\left(\epsilon,\mathcal{F}, \|\cdot\|_{\infty,n}\right)$, since $\|f\|_{2,n} \leq \|f\|_{\infty,n}$ for all $f \in \mathcal{F}$. Therefore, we obtain
\begin{equation}
\label{bound_cover_number_2}
    \log N\left(\epsilon,\mathcal{F}, \|\cdot\|_{2,n}\right) \leq \operatorname{Pdim}(\mathcal{F}) \cdot \log \left( \frac{2eB \cdot n}{\epsilon \cdot \operatorname{Pdim}(\mathcal{F})} \right) \leq \operatorname{Pdim}(\mathcal{F}) \cdot \log \left(\frac{2e B \cdot n}{\epsilon} \right),
\end{equation}
where the last inequality comes from $\operatorname{Pdim}(\mathcal{F}) \geq 1$.

Combining with \cref{bound_rademacher_complexity} and \cref{bound_cover_number_2}, we have
\begin{equation*}
    \begin{split}
        R_{S}(\mathcal{F}) & \leq 12 \int_{0}^{B} \sqrt{\frac{\operatorname{Pdim}(\mathcal{F})}{n} \cdot \log \left(\frac{2e B \cdot n}{\epsilon} \right) } d \epsilon \leq 12B \int_{0}^{1}\sqrt{\frac{\operatorname{Pdim}(\mathcal{F})}{n} \cdot \log \left(\frac{2e \cdot n}{\epsilon} \right) } d \epsilon\\
        & \leq 12B \int_{0}^{1}\sqrt{\frac{\operatorname{Pdim}(\mathcal{F})}{n}} \cdot \left(\sqrt{\log(2e) + \log n} + \sqrt{\log\left(\frac{1}{\epsilon}\right)}\right) d \epsilon\\
        & \leq 12B \int_{0}^{1}\sqrt{\frac{\operatorname{Pdim}(\mathcal{F})}{n}} \cdot \left(\sqrt{\log(2e) + \log n}  + 1 \right) d \epsilon\\
        & \leq 12B C_{0} \cdot \sqrt{\frac{\operatorname{Pdim}(\mathcal{F}) \cdot \log n}{n}},
    \end{split}
\end{equation*}
where the constant $C_{0}$ satisfies $\sqrt{\log(2e) + \log n}  + 1 \leq C_{0} \cdot \sqrt{\log n} $ and $C_{0} \approx 1$ if $\log n \gg 1$. 

Equipped with these results, we have
\begin{equation}
    \begin{split}
        \mathbb{E} I_1 & = \mathbb{E} \max_{f_{\alpha} \in \mathcal{F}_{\text{GS}}} \left \{ \mathbb{E}_{Z \sim \pi} f(g_{\hat{\theta}_{m,n}}(Z)) - \hat{\mathbb{E}}^m_{Z \sim \pi} f(g_{\hat{\theta}_{m,n}}(Z))\right \}\\
        & \leq \mathbb{E} \max_{f_{\alpha} \cdot  g_{\theta} \in \mathcal{F}_{\text{GS}} \circ \mathcal{G}} \left \{ \mathbb{E}_{Z \sim \pi} f(g_{\theta}(Z)) - \hat{\mathbb{E}}^m_{Z \sim \pi} f(g_{\theta}(Z))\right \} \leq 2 \mathcal{R}_{m}(\mathcal{F}_{\text{GS}} \circ \mathcal{G} )\\
        & \leq 24 C_{0} \cdot (L \cdot F(\pi,m) + M) \cdot \sqrt{\frac{\operatorname{Pdim}(\mathcal{F}_{\text{GS}} \circ \mathcal{G} ) \cdot \log m}{m}}.
    \end{split}
\end{equation}
Similarly, 
\begin{equation}
    \begin{split}
        \mathbb{E}I_2 & = \mathbb{E}\max_{f_{\alpha} \in \mathcal{F}} \left \{ \hat{\mathbb{E}}_{X \sim \nu}^n f(X) - \mathbb{E}_{X \sim \nu} f(X) \right \}  \leq 2 \mathcal{R}_{n}(\mathcal{F}_{\text{GS}})\\
        & \leq 24 C_{0} \cdot  F(\nu,n) \cdot \sqrt{\frac{\operatorname{Pdim}(\mathcal{F}_{\text{GS}}) \cdot \log n}{n}},
    \end{split}
\end{equation}
which completes the proof of \cref{new_generalization_thm}.

Note that if the source and the target distributions are bounded or sub-Gaussian, then  we have $F(\pi,m) \leq \mathcal{O}(\log m)$ and $F(\nu,n) \leq \mathcal{O}(\log n)$, and thus the generalization error bound is of $\widetilde{\mathcal{O}}(m^{-1/2})$ and $\widetilde{\mathcal{O}}(n^{-1/2})$ with respect to 
$m$ and $n$. 
Although the derived error bound is within our expectation, it is monotonically increasing with the pseudo-dimension of $\mathcal{F}_{\text{GS}}$ and $\mathcal{F}_{\text{GS}} \circ \mathcal{G}$, which are proportional to the width and depth of NNs \cite{anthony1999neural}. 
We remark that both \cref{generalization_thm} and 
\cref{generalization_thm2} provide upper bounds for the generalization of WGANs with GroupSort networks as their discriminators by optimal transport and machine 
learning approaches, respectively. 

\section{Experimental results}
\label{experiments}

Our theoretical analysis in Section 3 conveyed that with sufficient training data and properly chosen depth and width of generators and discriminators, WGANs can approximate target distributions in Wasserstein-1 distance. In this section, both synthetic (Swiss roll) and actual (MNIST) datasets are used to verify 
our theoretical results. Although we derive the upper bound for $\mathbf{Wass}_1(\mu_{\hat{\theta}_{m,n}},\nu)$, numerical experiments confirm the consistency between the upper bounds and the actual Wasserstein-1 distances, at least in the trend.

The estimated Wasserstein distances between the target distributions and the generated distributions were computed using the POT package \cite{flamary2021pot} on 10,000 samples. We repeated each experiment six times and then calculated the averaged Wasserstein distance to represent the expected value. For computational efficiency, we used mini-batch stochastic gradient descent rather than updating the whole dataset. Moreover, we generated noise at each iteration step because of its availability and cheapness. The features of our interest are the number of training data (denoted as $n$, consistent with those in theoretical analysis), as well as the width ($W_g$ and $W_f$, respectively) and depth ($D_g$ and $D_f$, respectively) of both the generator and the discriminator. A relaxed (proximal) constraint on the parameters was adopted \cite{anil2019sorting,tanielian2021approximating}. Similarly, we used the Bj\"{o}rck orthonormalization method to constrain parameters to be unit in $||\cdot||_2$ norm.  See \cite{bjorck1971iterative, anil2019sorting} for more details about Bj\"{o}rck orthonormalization.

\subsection{Synthetic dataset (Swiss roll)}
To show the approximation of WGANs to data lying on low-dimensional manifolds, in this subsection we use a two-dimensional Swiss roll (on a one-dimensional manifold) as synthetic data., with Swiss roll curves sampled from $
(x,y) \sim (z \cdot \text{cos} (4 \pi z), z \cdot \text{sin} (4 \pi z)); \hspace{0.5em} z \sim \text{Uniform}([0.25,1])
$.
\begin{figure}[h]
  \centering
  \subfloat[1000 training data] {
     \label{syn_num_train:a}     
    \includegraphics[width=0.3\textwidth]{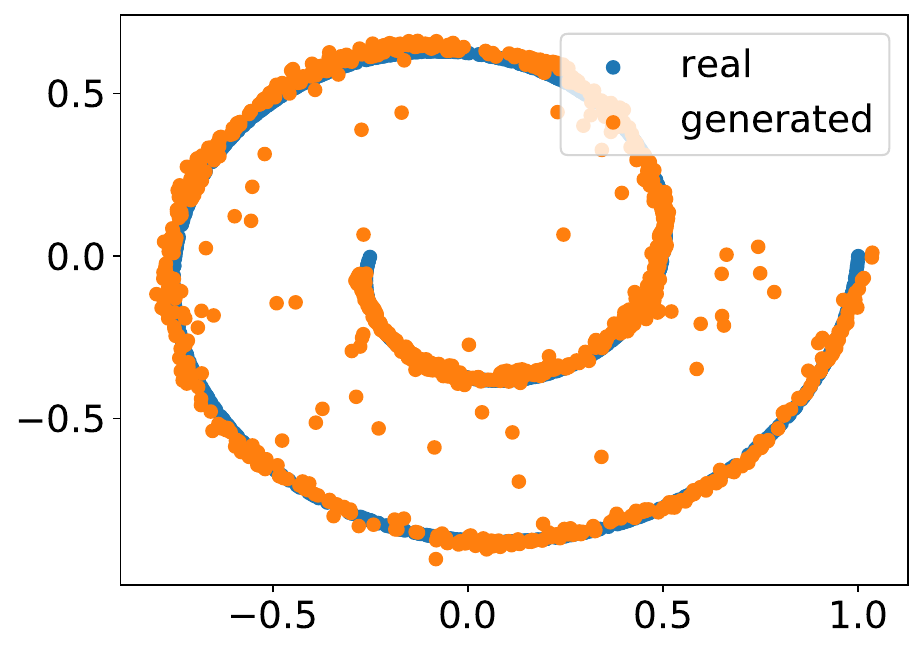}
    %\resizebox{0.31\textwidth}{!}{\input{fig/train_num_1000.pdf}}
    }\   
    \subfloat[6000 training data] {
    \label{syn_num_train:b}     
    \includegraphics[width=0.3\textwidth]{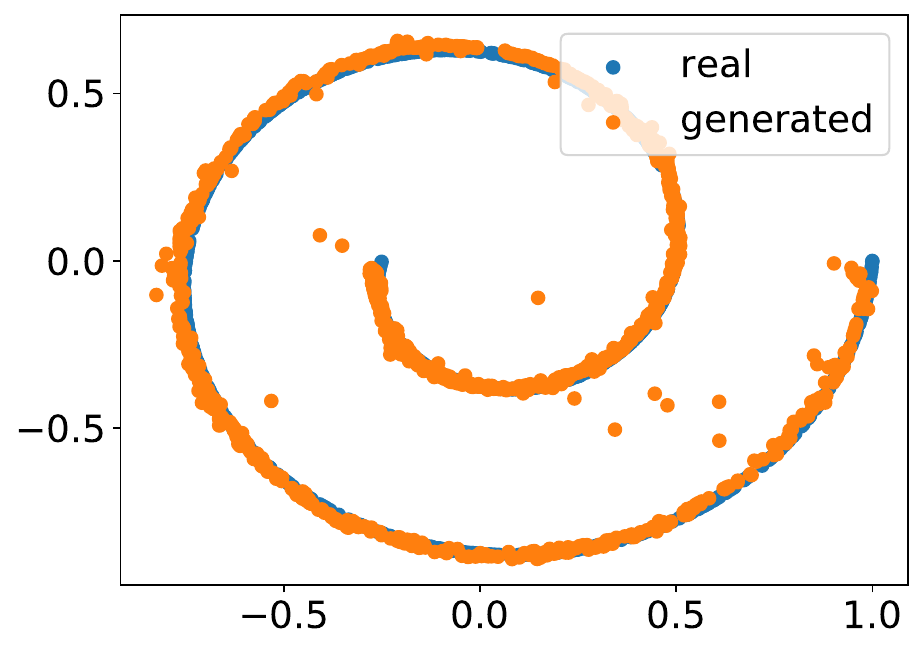}
    %\resizebox{0.31\textwidth}{!}{\input{fig/train_num_6000.pgf}}
    }\
    \subfloat[number of training data]{
    \label{syn_num_train:c}   
    %\resizebox{0.31\textwidth}{!}{\input{fig/syn_numtrain.pgf}}
    \includegraphics[width=0.31\textwidth]{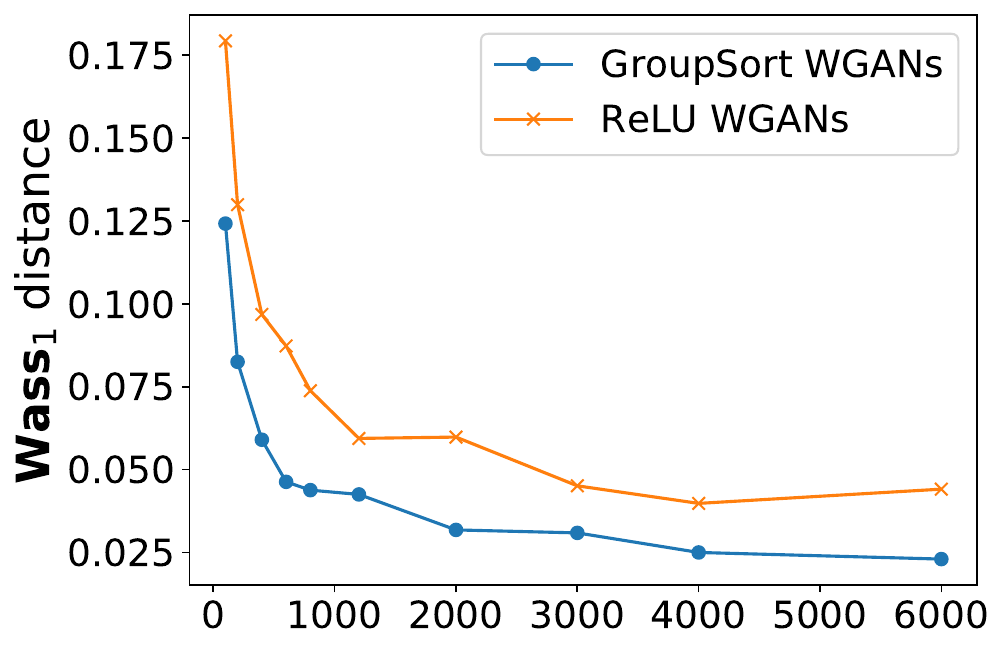}
    }\
    \caption{(GroupSort NNs as discriminators)
%		Numerical results for fixed models but different numbers of training data. %The architectures of the generator and the discriminator are fixed. Both the %generator and the discriminator are of width 30 and depth 2. 		
		%(a), (b): 
		Visualized real data and generated data with (a) 1000 and 
 		(b) 6000 samples for training respectively; (c) The $\mathbf{Wass}_1$ distance between the generated distribution and the real distribution with different numbers of training data. Both the generator and the discriminator are fixed in their width (30) and depth (2).}
    \label{syn_num_train}
\end{figure}

To improve the convergence and stability of the generator and the discriminator, they were trained by Adam with default hyperparameters 
(listed in \cref{tab:table-hyperprm} in \cref{append_hypm}) for 50,000 iterations. For each iteration, we first updated the discriminator five times and then updated the generator only once.
\cref{syn_num_train} shows the relationship between the approximation of the generated distribution and the number of training data. When trained on 1000 synthetic samples, the generator can generate fake data that are sufficiently close to the real data. Moreover, with more training data (e.g., 6000 samples), the generated distribution approximates the target distribution better, with hardly any generated points straying from the Swiss roll curve. The curve in \cref{syn_num_train:c} suggests that the approximation of the generated distribution improves with more training data. When $n=100$, the training is unstable and fails 
% because it violates our assumption that $W_g \lesssim m^{1/(6rD_g)}$. A more straightforward explanation is that the excessive flexibility of our model (the number of parameters to be estimated exceeds the number of training data). 
because in this case $\mathbf{Wass}_1(\hat{\nu},\nu)$ dominates the error, and its variance (of  $O(\frac{1}{n})$ \cite{del2019central, bernton2019parameter}) causes the numerical results to be unstable.
When $n \geq 2000$, the curve for the estimated $\mathbf{Wass}_1$ distance moves slowly because the dominant terms of the error come from the generator and the discriminator. \cref{syn_num_train} also shows that using GroupSort NNs outperforms using ReLU NNs in WGANs.

\begin{figure}
%[H]
% \makebox[\linewidth][c]{%
  \centering
  \subfloat
	[(GroupSort) width of the generator] 
	{
     \label{syn_g_d:a}     
    %\resizebox{0.28\textwidth}{!}{\input{fig/syn_g_width.pgf}}
    \includegraphics[width=0.35\textwidth]{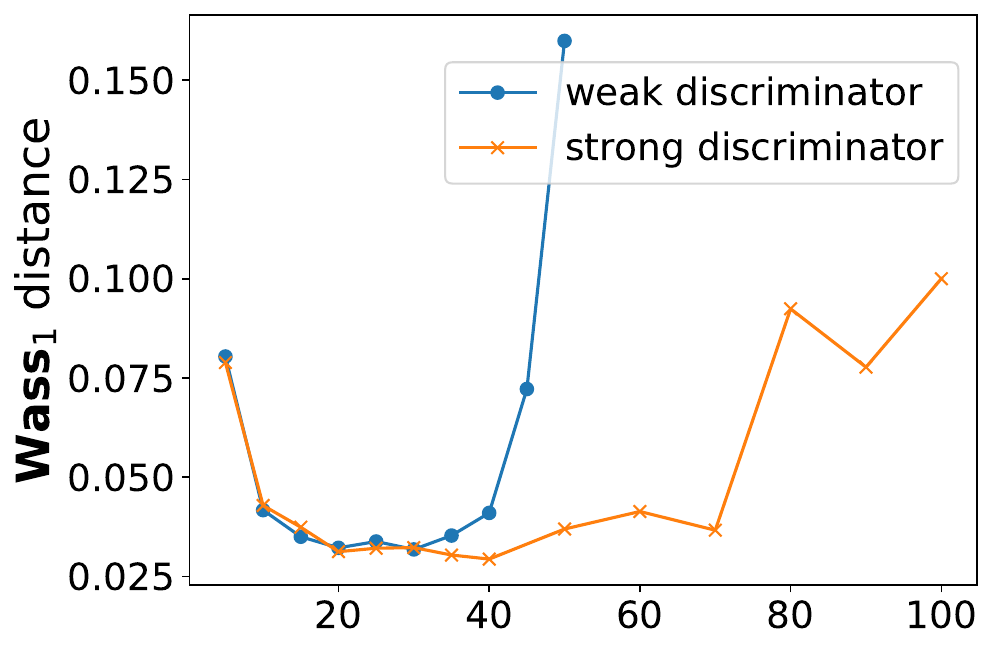}
    }\  
    \subfloat
  [(ReLU) width of the generator ] 
  {
     \label{syn_relu_g_d:b}     
    %\resizebox{0.28\textwidth}{!}{\input{fig/syn_relu_g_width.pgf}}
    \includegraphics[width=0.35\textwidth]{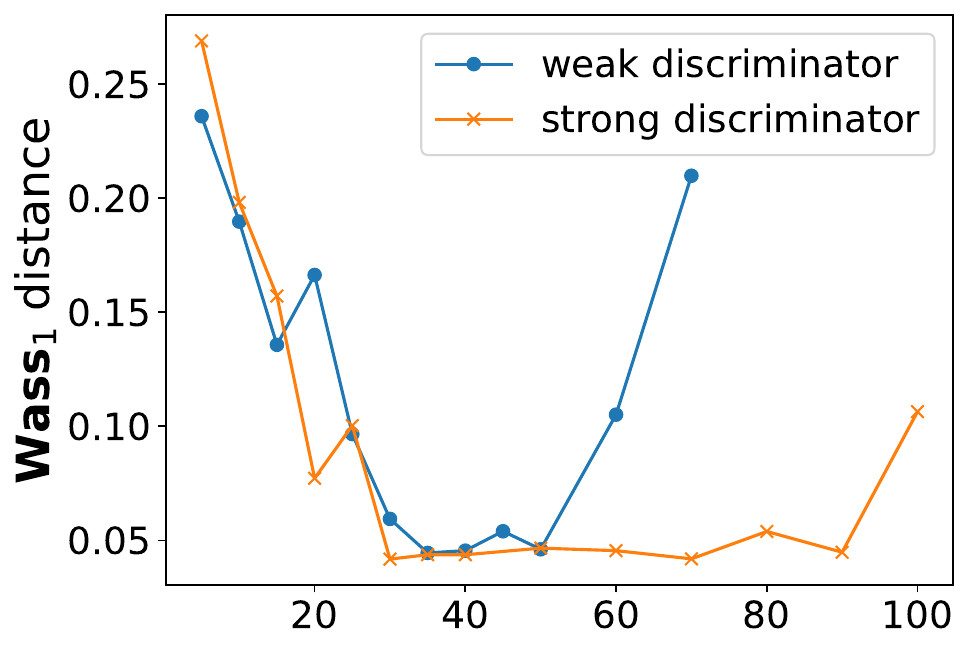}
    }\\
    \subfloat
	[(GroupSort) depth of the generator] 
	{
    \label{syn_g_d:b}     
    %\resizebox{0.28\textwidth}{!}{\input{fig/syn_g_depth.pgf}}
    \includegraphics[width=0.35\textwidth]{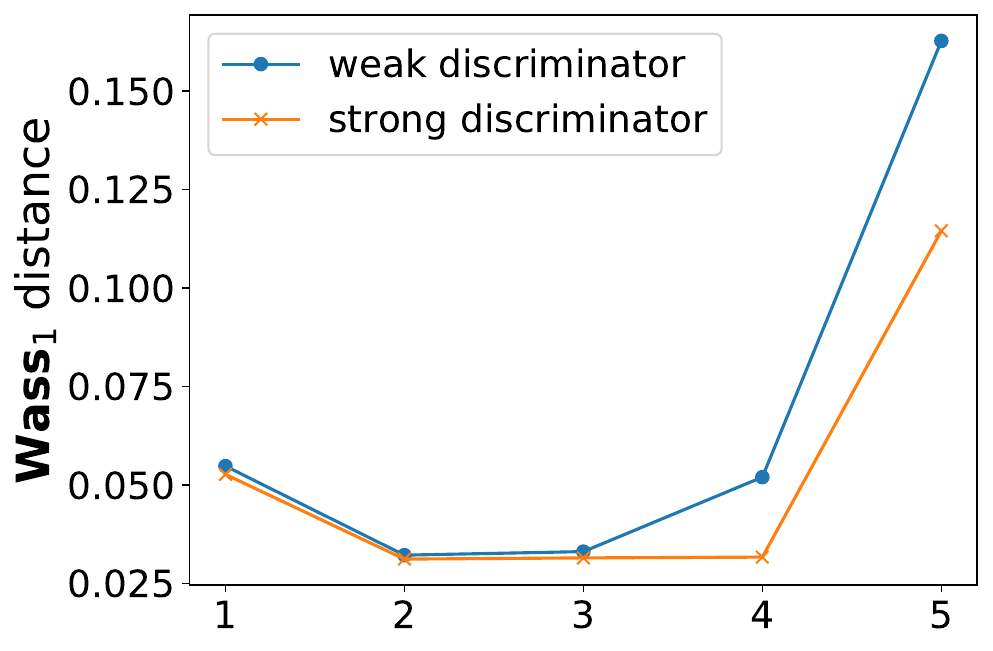}
    }\
    \subfloat
    [(ReLU) depth of the generator] 
    {
    \label{syn_relu_g_d:c}     
    %\resizebox{0.28\textwidth}{!}{\input{fig/syn_relu_g_depth.pgf}}
    \includegraphics[width=0.35\textwidth]{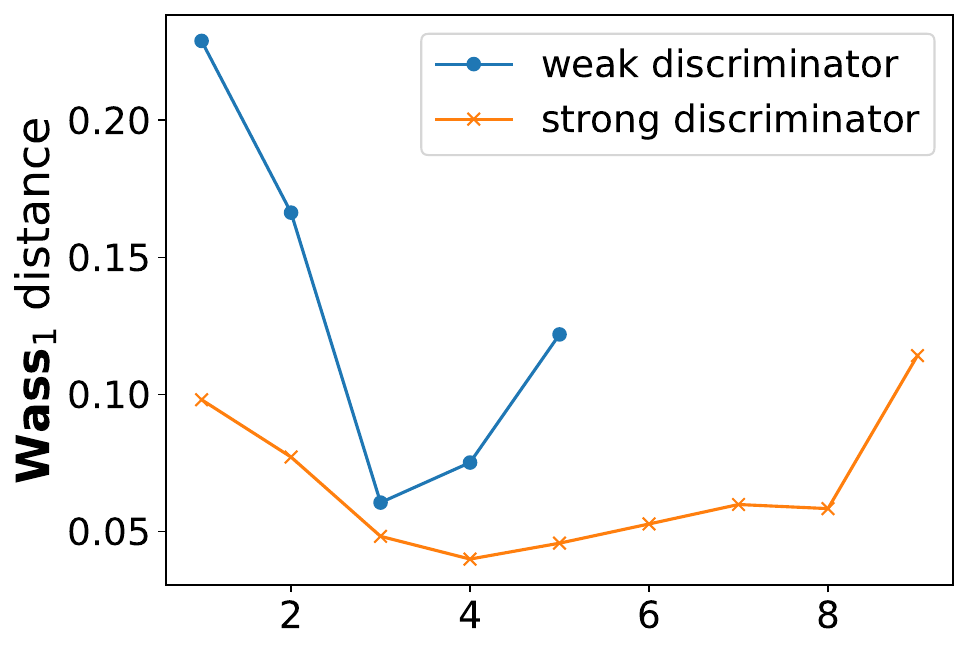}
    }
    \\ 
    \subfloat
	[(GroupSort) width of the discriminator] 
	{
    \label{syn_g_d:c}     
    %\resizebox{0.28\textwidth}{!}{\input{fig/syn_d_width.pgf}}
    \includegraphics[width=0.35\textwidth]{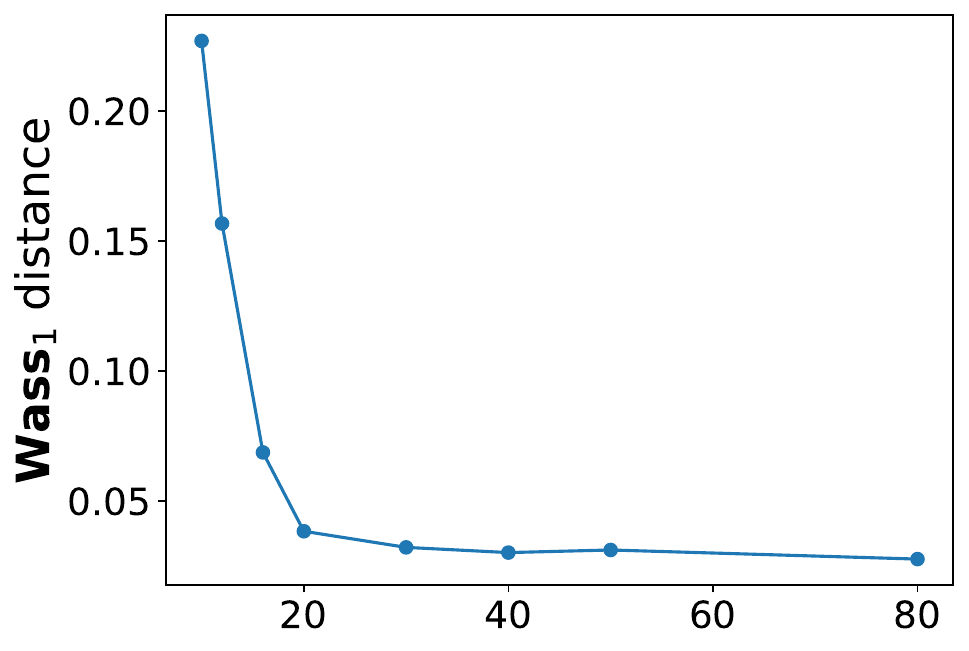}
    }\
     \subfloat
    [(ReLU) width of the discriminator] 
    {
    \label{syn_relu_g_d:d}     
    %\resizebox{0.28\textwidth}{!}{\input{fig/syn_relu_d_width.pgf}}
    \includegraphics[width=0.35\textwidth]{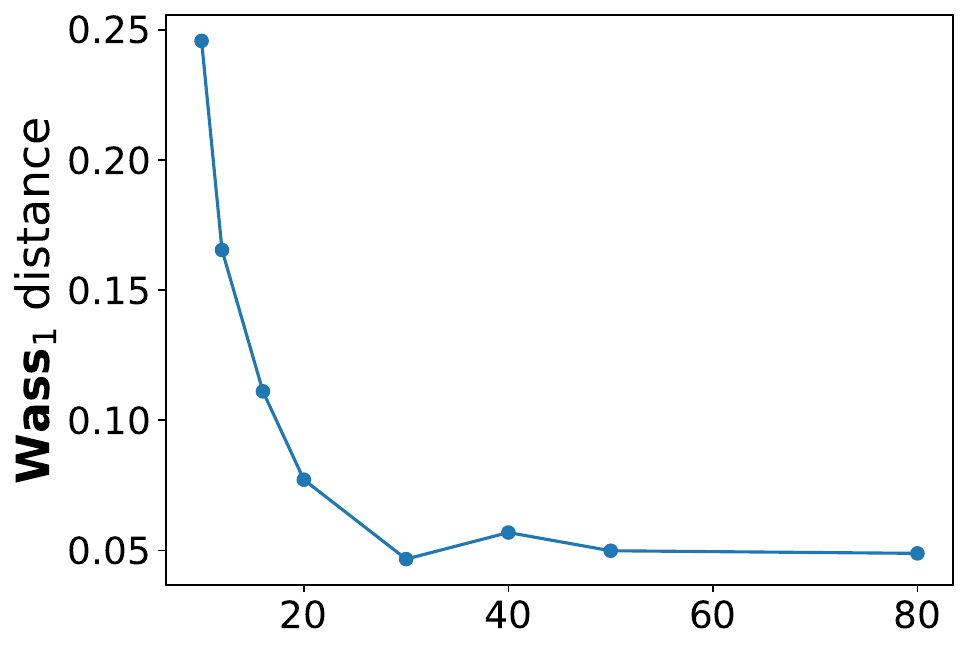}
    }\\
    \subfloat
	[(GroupSort) depth of the discriminator] 
	{
    \label{syn_g_d:d}     
    %\resizebox{0.28\textwidth}{!}{\input{fig/syn_d_depth.pgf}}
    \includegraphics[width=0.35\textwidth]{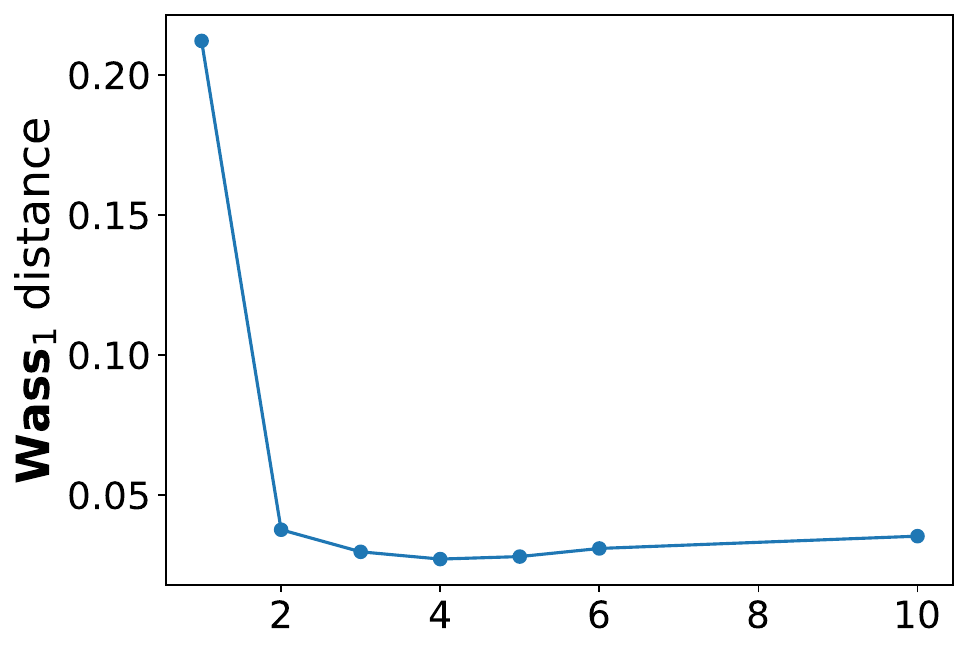}
    }\
    \subfloat
    [(ReLU) depth of the discriminator] 
    {
    \label{syn_relu_g_d:e}     
    %\resizebox{0.28\textwidth}{!}{\input{fig/syn_relu_d_depth.pgf}}
    \includegraphics[width=0.35\textwidth]{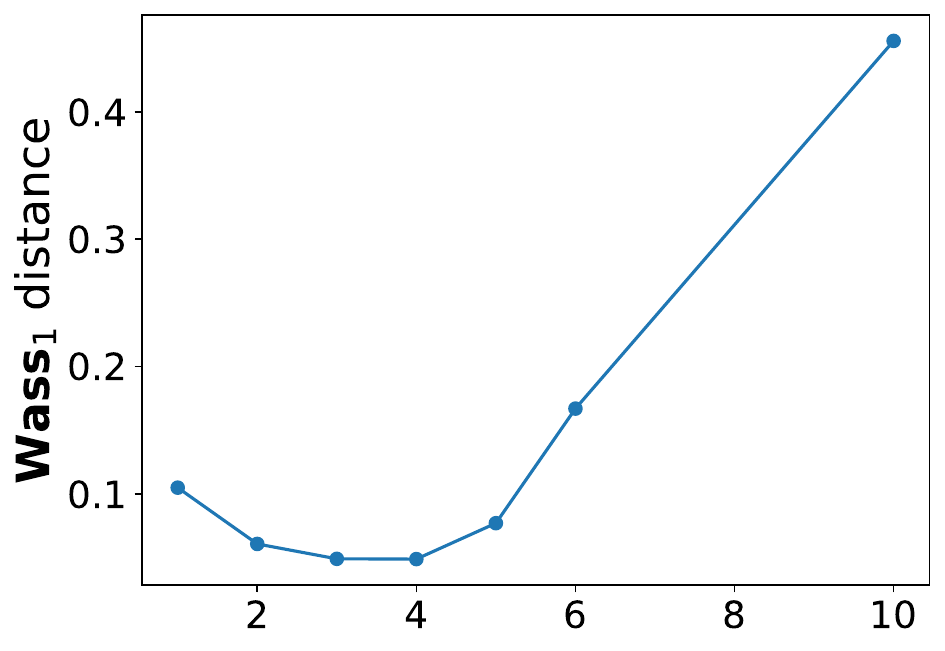}
    }\
%     }
    \caption{\textbf{First column}: GroupSort NN as a discriminator.
		%Numerical results for the same number of training data (2000 swiss roll data) but different architectures of models. The three figures display the estimated $\mathbf{W}_1$ distances between the generated distributions and the real distributions. 
		$\mathbf{Wass}_1$ distance 
%		between the generated distributions and the real distributions 
with respect to   	
		(a) $W_g$ using a strong
		discriminator $(D_f,W_f)=(50,2)$ and a weak
		discriminator $(D_f,W_f)=(30,2)$;
		(c) $D_g$ using a strong
		discriminator $(D_f,W_f)=(50,2)$ and a weak
		discriminator $(D_f,W_f)=(30,2)$;
		(e) $W_f$ using a generator $(D_g,W_g)=(20,2)$;
	    (g) $D_f$ using a generator $(D_g,W_g)=(20,2)$.
     \textbf{Second column}: ReLU NN as a discriminator.	The $\mathbf{Wass}_1$ distance 
     %between the generated distribution and the real distribution 
     with respect to (a) 
		different numbers of training data;
		(b) $W_g$ using a strong
		discriminator $(D_f,W_f)=(50,2)$ and a weak
		discriminator $(D_f,W_f)=(30,2)$;
		(d) $D_g$ using a strong
		discriminator $(D_f,W_f)=(50,2)$ and a weak
		discriminator $(D_f,W_f)=(30,2)$;
		(f) $W_f$ using a generator $(D_g,W_g)=(20,2)$;
	  (h) $D_f$ using a generator $(D_g,W_g)=(20,2)$.}  	
    \label{syn_g_d}
\end{figure}

In the first (i.e., left-hand) column of \cref{syn_g_d}, we investigate 
how the width and depth of the generator and GroupSort discriminator affect the learning of the target distribution. Here, the number of training data 
is fixed at $n=2000$.
\cref{syn_g_d:a} shows that with increasing generator width, the $\mathbf{Wass}_1$ distance decreases initially,
then remains approximately constant, and then increases; 
therefore, the generator capacity cannot improve the approximation results when the dominant error comes from the discriminator capacity and the limited number of training data. 
Interestingly, note that the $\mathbf{Wass}_1$ distance increases significantly
when a weak discriminator is used. 
This also illustrates the importance and high requirements of the discriminator in GANs. 
\cref{syn_g_d:b} shows
similar results, i.e., that overly deep generators will cause worse approximation. 
\cref{syn_g_d:a} and \cref{syn_g_d:b} combined
confirm our conclusion that overly deep and wide (high-capacity) generators may cause worse results (after training) than do low-capacity generators if the discriminators are insufficiently strong.
In \cref{syn_g_d:c} and \cref{syn_g_d:d}, we 
fix a generator with $(W_g,D_g)=(20,2)$.
Clearly, increasing discriminator capacity improves the approximation of WGANs, which is consistent with the trend of our theoretical error-bound results.

On the other hand, the second (i.e., right-hand) column of \cref{syn_g_d} shows the experimental results for ReLU WGANs with clipping (i.e., the original WGAN). 
The experimental settings are the 
same as those in \cref{syn_num_train} and the first column of \cref{syn_g_d}, and the hyperparameters are listed in \cref{tab:table-hyperprm-relu} of \cref{append_hypm}. 
\cref{syn_g_d} shows that the results are similar 
for GroupSort-WGANs and ReLU-WGANs. 
From \cref{syn_num_train:c} and \cref{syn_g_d},
we make
the following summary observations.
% from \cref{syn_num_train:c} and \cref{syn_g_d}.
\begin{enumerate}
\item[(i)] The $\mathbf{Wass}_1$ distance obtained by using GroupSort NNs is 
smaller than that by using ReLU NNs.
\item[(ii)] Similar to the results in \cref{syn_g_d:a}, 
the generator capacity cannot improve
the approximation results, and 
overly deep and wide (high-capacity) generators cause worse results (after training)
than do low-capacity generators if discriminators are insufficiently strong.
\item[(iii)] A striking difference lies in the results of varying the discriminator depth. Deep ReLU discriminators lead to training failure, which can be explained by the results in \cite{yarotsky2017error} that although ReLU feed-forward NNs can approximate any 1-Lipschitz functions, their parameters cannot be constrained. More precisely, the clipping operation that constrains the norm of the parameters breaks the capacity of the ReLU discriminator, whose activation functions do not preserve norms. 
This is the main reason why the convergence and generalization of WGANs with ReLU activation functions have not been shown and studied before. 
The main contribution herein is to overcome this problem and establish 
generalization and convergence properties of WGANs with 
GroupSort NNs under mild conditions. 
\end{enumerate}

\subsection{MNIST dataset}

Although there are no explicit probability distributions defined on MNIST data, empirical results show consistency with our theory.
% and the practice on 
%MNIST dataset.
%) with implicit distributions. 
The MNIST data consisting of 60,000 images 
are defined on [-1,1] after scaling, meaning that they come from a bounded distribution. Therefore, \cref{decay} for the target distribution is automatically satisfied. For image data, we apply the tanh activation function to the output layer for its ability to monotonically map real numbers to [-1, 1].  The selections of hyperparameters are listed in \cref{tab:mnist_hyperprm} of 
\cref{append_hypm}. We use principal component analysis (PCA) to reduce the dimension of the image data (to 50 dimensions) and then calculate the $\mathbf{Wass}_1$ distance.

\begin{figure}[h]
% \makebox[\linewidth][c]{%
  \centering
  \subfloat
  [width of the generator]
  {
     \label{mnist_g_d:a}     
    %\resizebox{0.28\textwidth}{!}{\input{fig/mnist_g_width.pgf}}
    \includegraphics[width=0.35\textwidth]{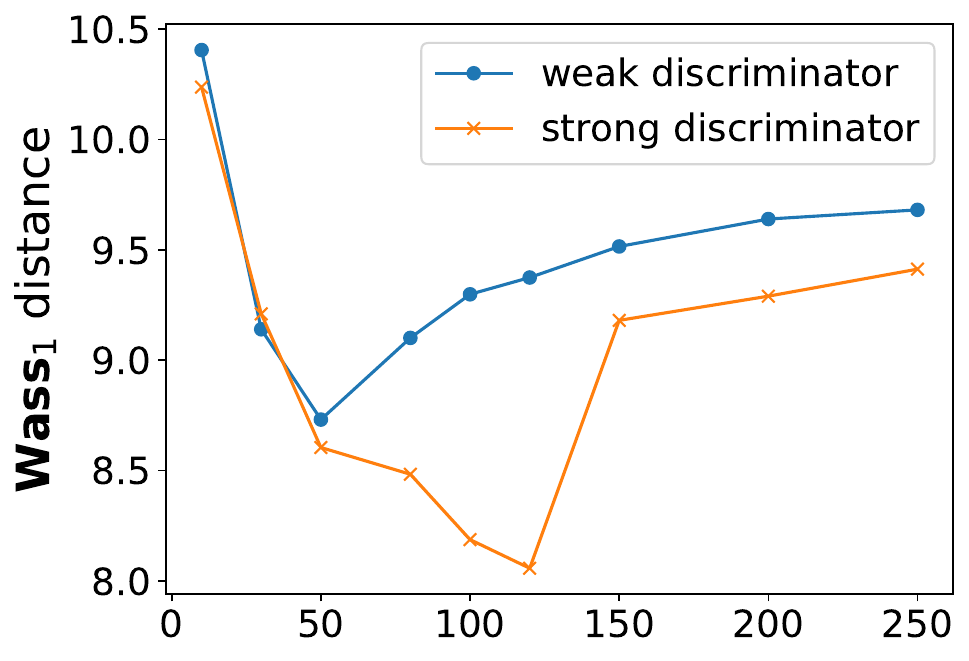}
    }\  
    \subfloat
    [depth of the generator] 
    {
    \label{mnist_g_d:b}     
    %\resizebox{0.28\textwidth}{!}{\input{fig/mnist_g_depth.pgf}}
    \includegraphics[width=0.35\textwidth]{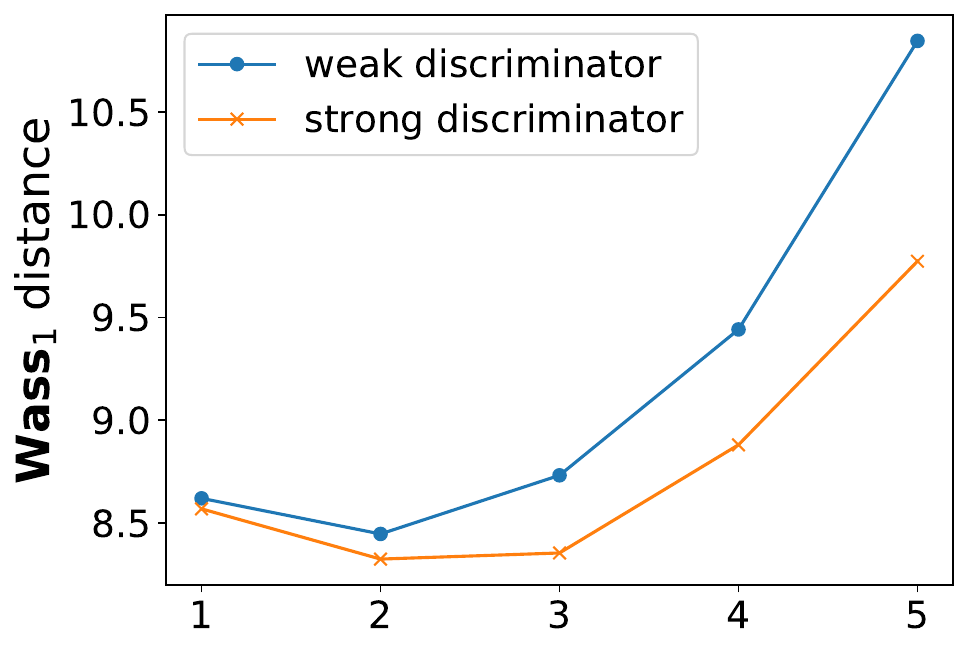}
    }\\
    \subfloat
    [width of the discriminator] 
    {
    \label{mnist_g_d:c}     
    %\resizebox{0.28\textwidth}{!}{\input{fig/mnist_d_width.pgf}}
    \includegraphics[width=0.35\textwidth]{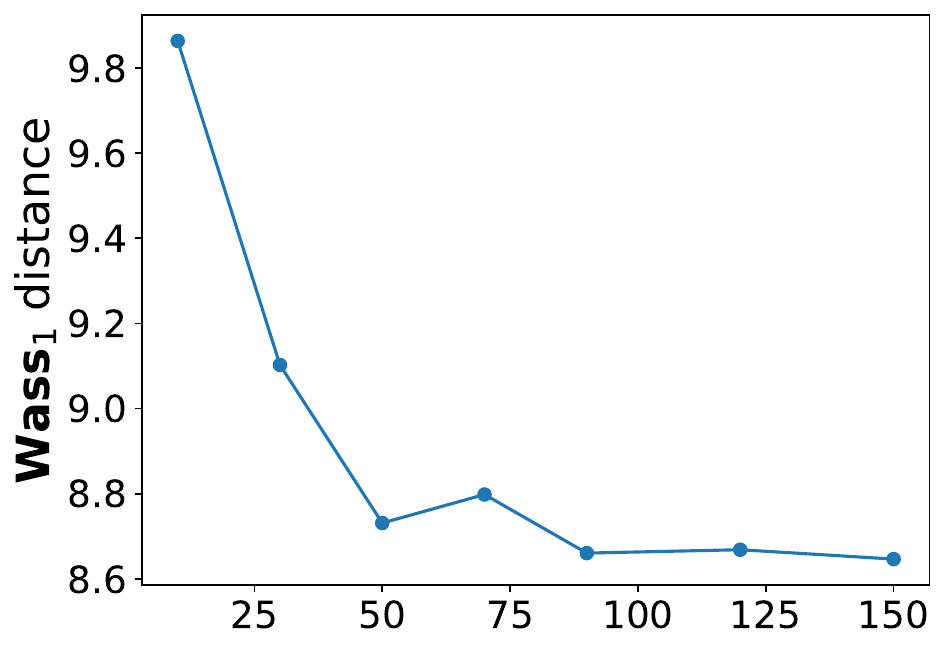}
    }\
    \subfloat
    [depth of the discriminator] 
    {
    \label{mnist_g_d:d}     
    %\resizebox{0.28\textwidth}{!}{\input{fig/mnist_d_depth.pgf}}
    \includegraphics[width=0.35\textwidth]{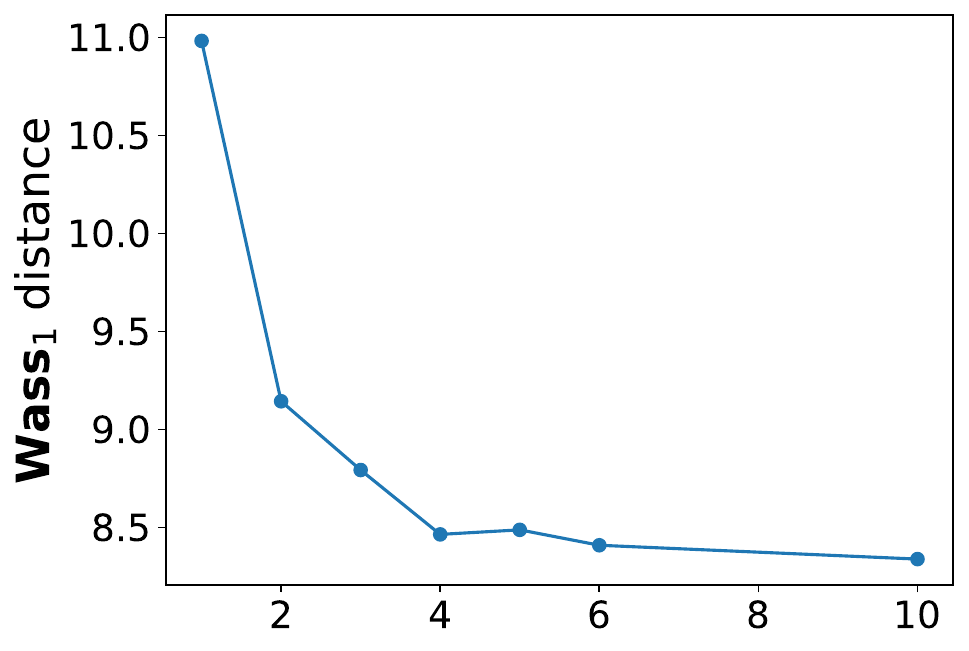}
    }\
%     }
    \caption{
  (GroupSort NN as a discriminator)
		The $\mathbf{Wass}_1$ distance between the generated distributions and the real distributions with respect to   		
		(a) $W_g$ using a strong
		discriminator $(D_f,W_f)=(100,3)$ and a weak
		discriminator $(D_f,W_f)=(50,3)$;
		(b) $D_g$ using a strong
		discriminator $(D_f,W_f)=(50,3)$ and a weak
		discriminator $(D_f,W_f)=(100,4)$;
		(c) $W_f$ using a generator $(D_g,W_g)=(50,3)$;
	    (d) $D_f$ using a generator $(D_g,W_g)=(50,3)$.}  
    \label{mnist_g_d}
\end{figure}

As shown in \cref{mnist_g_d}, the numerical results on the MNIST dataset are similar to those on the synthetic dataset (Swiss roll). 
% However, there are some differences between Figur \ref{syn_g_d:a} and Figure \ref{MNIST_g_d:a}. 
The rise of $\mathbf{Wass}_1$ distance with increasing $W_g$ and $D_g$ in \cref{mnist_g_d:a} and \cref{mnist_g_d:b} 
is as a consequence of the fact that the generator capacity (i.e., $L=(W_g \cdot \Gamma)^{D_g}$) dominates the error $L \cdot (m^{-1/2}\log m \vee m^{-1/r})$ and $ L \cdot ( 2^{-D_f/(2d^2)} \vee W_f^{-1/(2d^2)})$. 
This implies that 
overly deep and wide (high-capacity) generators cause worse results (after training)
than do low-capacity generators, if the discriminators are not strong enough. 
Moreover, the case with a strong discriminator always has a lower error than the case with a weak discriminator. The loss of strong discriminators rises up later than the loss of weaker discriminators. They are consistent with our theoretical results in (\cref{main_thm}). 
In \cref{mnist_g_d:c} and \cref{mnist_g_d:d}, we 
fix a generator with $(W_g,D_g)=(50,3)$.
Clearly, increasing the discriminator capacity improves the 
approximation of WGANs, which is consistent with the trend of 
our theoretical error-bound results.
For illustration, 
some visualized results are shown in \cref{append_vis}. 

\section{Conclusion}
\label{conclusion}
Herein, we presented an error bound in Wasserstein distance for the approximation of WGANs (with GroupSort NNs as discriminators). We showed that with sufficiently many training samples, for generators and discriminators with properly chosen width and depth (satisfying the conditions in \cref{main_thm}), the GroupSort WGANs can approximate distributions well. The error comes from (i) the finite number of samples, (ii) the capacity of generators, and (iii) the approximation of discriminators to 1-Lipschitz functions. It shows that there are strict requirements for the capacity of discriminators, especially for their width. More importantly, overly deep and wide (high-capacity) generators may cause worse results than those with low-capacity generators if the discriminators are insufficiently strong. The numerical results presented in Section 4 further confirmed our theoretical analysis. 
Moreover, we further developed a generalization error bound that is free from the curse of dimensionality with respect to the numbers of training data. 
%It reduces from $\widetilde{\mathcal{O}}(m^{-1/r} + n^{-1/d})$ to $
%\widetilde{\mathcal{O}}(\text{Pdim}(\mathcal{F}\circ \mathcal{G}) \cdot m^{-1/%2} + \text{Pdim}(\mathcal{F}) \cdot n^{-1/2})$. However, the latter seems to %contradict numerical observations. 
The theoretical results also provided some numerical insights: regardless of the available computation power, large numbers of training data, strong discriminators and generators are desired when training GANs. 
In future research, it would be interesting to study a tight upper bound for the generalization error of WGANs and analyze
proper regularizations for generators to improve the performances of GANs guided by theories.

\appendix

% \newpage
\section{Grouping size of GroupSort neural networks}
\label{grouping_size}
Here, we extend Theorem 2 and 3 in \cite{tanielian2021approximating} to \cref{approx_gs_nn} to theoretically show that grouping size contributes to the approximation capabilities of GroupSort NNs. 
\begin{theorem}
[Approximation of GroupSort neural networks to 1-Lipschitz functions]
\label{approx_gs_nn}
For any 1-Lipschitz function $f$ defined on $[-M_u,M_u]^{d}$, there exists $f_{\alpha} \in \mathcal{F}_{\text{GS}}$ with grouping size $k$, $D_f=n+1$ and $W_f=k^n$, such that $\|f_{\alpha}-f\|_{L^{\infty}([-M_u,M_u]^{d})} \leq 2M_u \cdot \epsilon$, where $\epsilon = 2\sqrt{d} \cdot  W_f^{-1/d^2}$. Furthermore, let $k=\sqrt{W_f}$ and $D_f = 3$, then $\|f_{\alpha}-f\|_{L^{\infty}([-M_u,M_u]^{d})} \leq 2M_u \cdot \epsilon$.
\end{theorem}
Note that the error $\epsilon =\mathcal{O}(W_f^{-1/d^2})$ with grouping size $k=\sqrt{W_f}$ and $\epsilon =\mathcal{O}(2^{-D_f/d^2} \vee W_f^{-1/d^2})$ with $k=2$ theoretically imply that large grouping size cannot reduce the error to be lower than $\mathcal{O}(W_f^{-1/d^2})$ for deep GroupSort NNs. Although theoretical results recommend using larger grouping sizes in GroupSort NNs, empirical results in \cite{anil2019sorting} show slight improvement of GroupSort NNs with larger grouping sizes, and they outperform ReLU NNs by a large margin.  GroupSort NNs work well with a grouping size of $2$, and the grouping size is not that important empirically. Moreover, it is more convenient to fix the grouping size to be $2$ because the width $W_f$ must be a multiple of the grouping size.

\section{Hyperparameters for numerical experiments} Here, we detailed the hyperparameters used in numerical experiments, as listed in \cref{append_hypm}, \cref{tab:table-hyperprm-relu} and \cref{tab:mnist_hyperprm}. 
\label{append_hypm}
\begin{table}[h]
\small
\caption{Hyperparameters used for training (GroupSort) WGANs on the synthetic dataset (Swiss Roll).}
\label{tab:table-hyperprm}
\centering
\begin{tabular}{ c|c } 
\hline
Operation & Features  \\
\hline
 Generator & \multirow{2}{*}{Architecture changed in each experiment} \\ 
 Discriminator &  \\ 
\hline
 Bj\"{o}rck iteration steps & 5\\
 Bj\"{o}rck order & 2 \\
 \hline
 Optimizer & Adam: $\beta_1 = 0.9$, $\beta_2 = 0.99$ \\ 
 Learning rate (for both generator and &  \multirow{2}{*}{1E-04}\\
 discriminator) & \\
 Batch size & 100 \\
 Noise &  Two-dimensional standard Gaussian noise,  $\mathcal{N}(0,I_2)$\\
 \hline
\end{tabular}
\end{table}

\begin{table}[h]
\caption{Hyperparameters used for training (ReLU) WGANs on the synthetic dataset (Swiss roll).}
\label{tab:table-hyperprm-relu}
\small
\centering
\begin{tabular}{ c|c } 
\hline
Operation & Features  \\
\hline
 Generator & \multirow{2}{*}{Architecture changed in each experiment} \\ 
 Discriminator &  \\ 
\hline
 Optimizer & RMSProp: $\rho=0.9$ \\ 
 Learning rate (for both generator and &  \multirow{2}{*}{5E-05}\\
  discriminator) & \\
 Batch size & 100 \\
 Noise &  Two-dimensional standard Gaussian noise,  $\mathcal{N}(0,I_2)$\\
 \hline
\end{tabular}
\end{table}

\begin{table}[h]
\caption{Hyperparameters used for training (GroupSort) WGANs on the actual dataset (MNIST).}
\label{tab:mnist_hyperprm}
\small
\centering
\begin{tabular}{ c|c } 
\hline
Operation & Features  \\
\hline
 Generator & \multirow{2}{*}{Architecture changed in each experiment} \\ 
 Discriminator &  \\ 
\hline
 Bj\"{o}rck iteration steps & 5\\
 Bj\"{o}rck order & 2 \\
 \hline
 Optimizer & Adam: $\beta_1 = 0.5$, $\beta_2 = 0.99$ \\ 
 Learning rate (for generator) &  5E-04\\
 Learning rate (for discriminator) & 1E-03\\
 Batch size & 512 \\
 Noise &  50-dimensional standard Gaussian noise,  $\mathcal{N}(0,I_{50})$\\
 \hline
\end{tabular}
\end{table}

\section{Visualized numerical results}
\label{append_vis}
\begin{figure}[htb]
  \centering
  \subfloat[$(D_g, W_g)=(3, 10)$, $(D_f, W_f)=(3, 50)$;\\ \centering  $\mathbf{Wass}_1=10.40$.] {
     \label{vis1:a}     
    \includegraphics[width=0.3\textwidth]{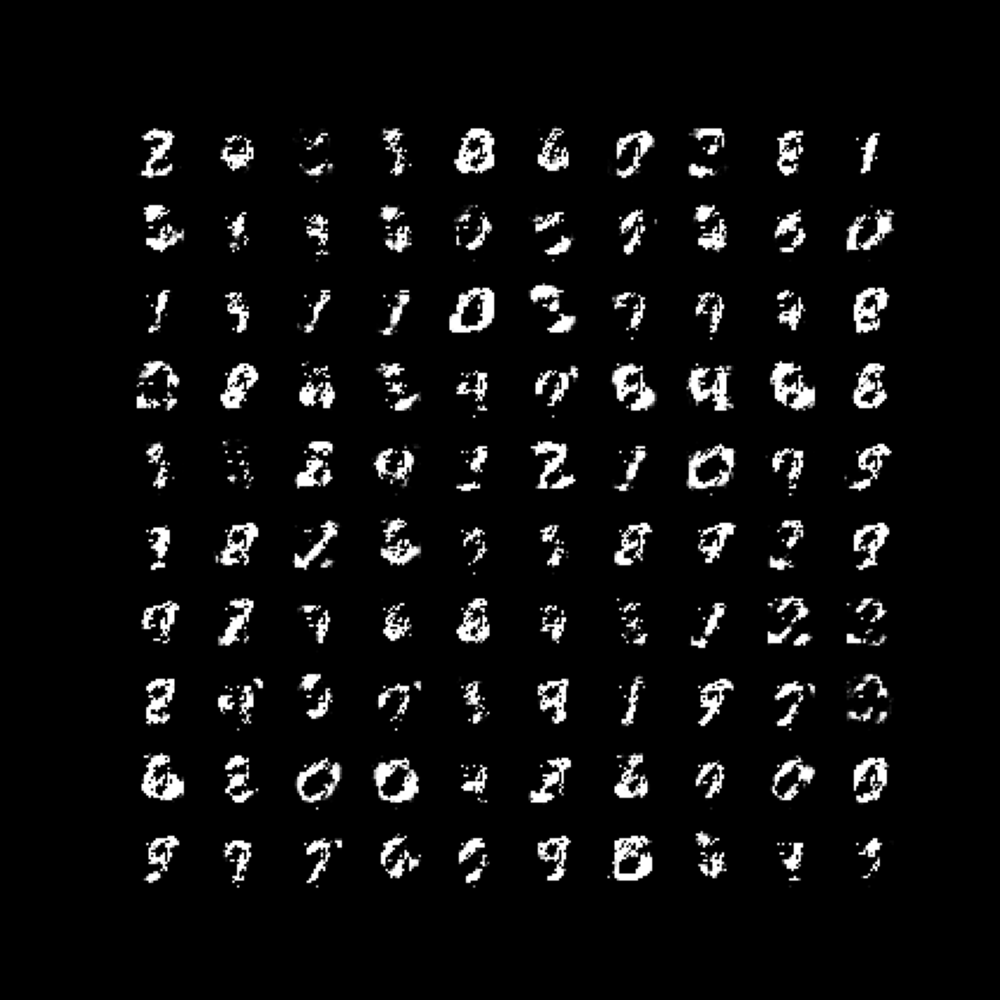}
    }\   
    \subfloat[$(D_g, W_g)=(3, 50)$, $(D_f, W_f)=(3, 50)$;\\ \centering $\mathbf{Wass}_1=8.73$.] {
    \label{vis1:b}     
    \includegraphics[width=0.3\textwidth]{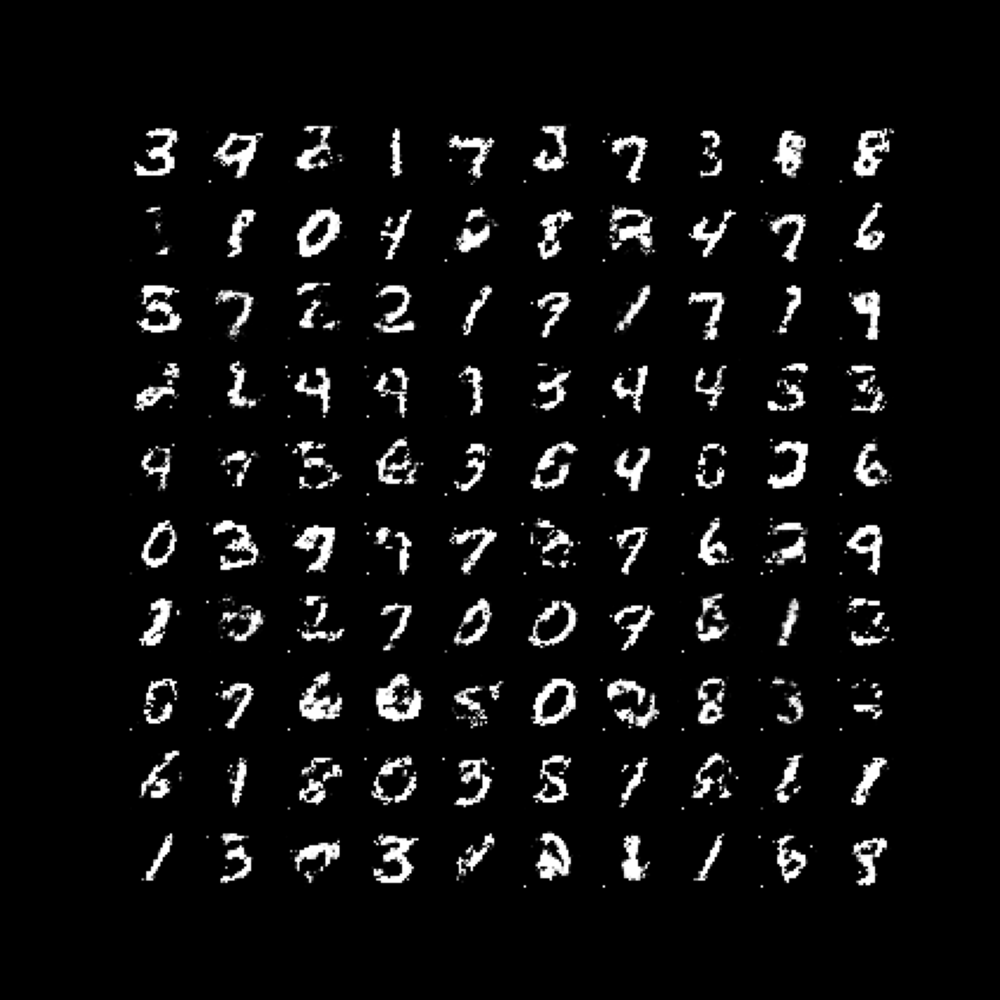}
    }\
    \subfloat[$(D_g, W_g)=(3, 200)$, $(D_f, W_f)=(3, 50)$;\\ \centering $\mathbf{Wass}_1=9.64$.] {
    \label{vis1:c}     
    \includegraphics[width=0.3\textwidth]{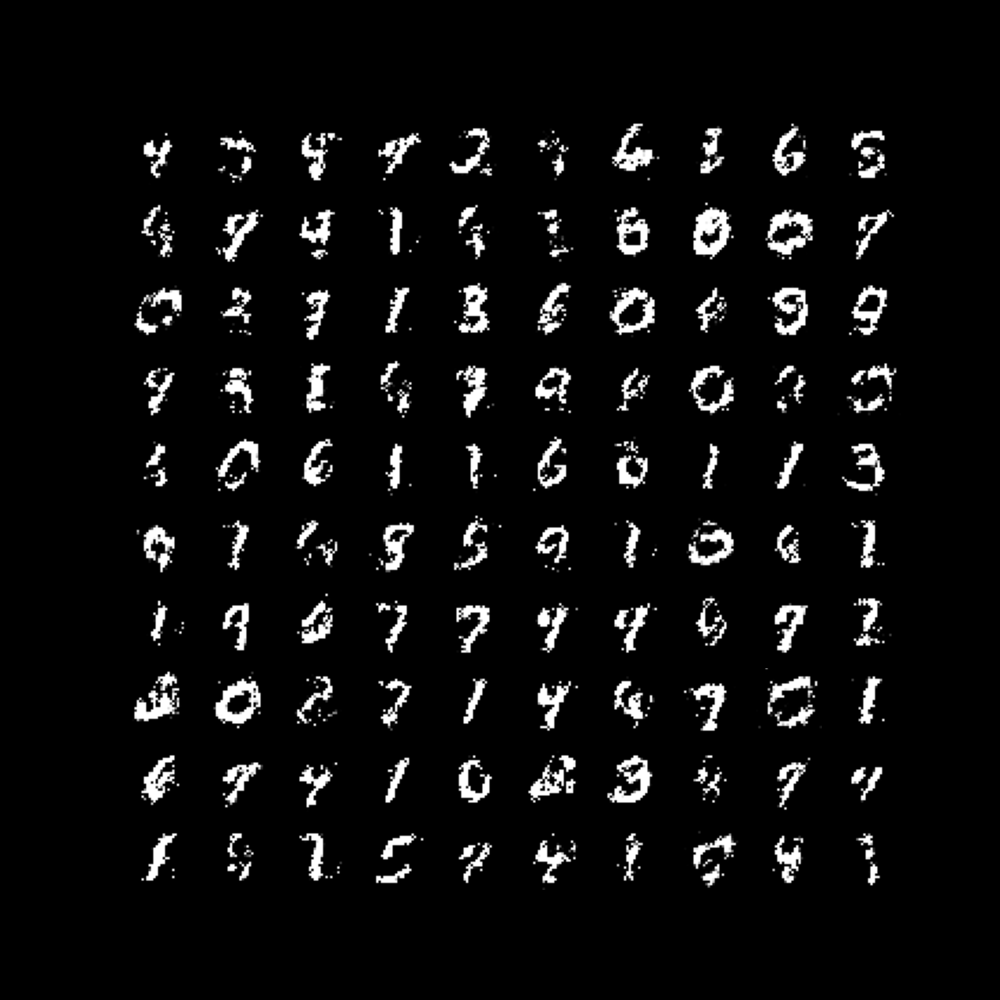}
    }\\
    \subfloat[$(D_g, W_g)=(3, 120)$, $(D_f, W_f)=(3, 100)$;\\ \centering  $\mathbf{Wass}_1=8.06$.] {
    \label{vis1:d}     
    \includegraphics[width=0.3\textwidth]{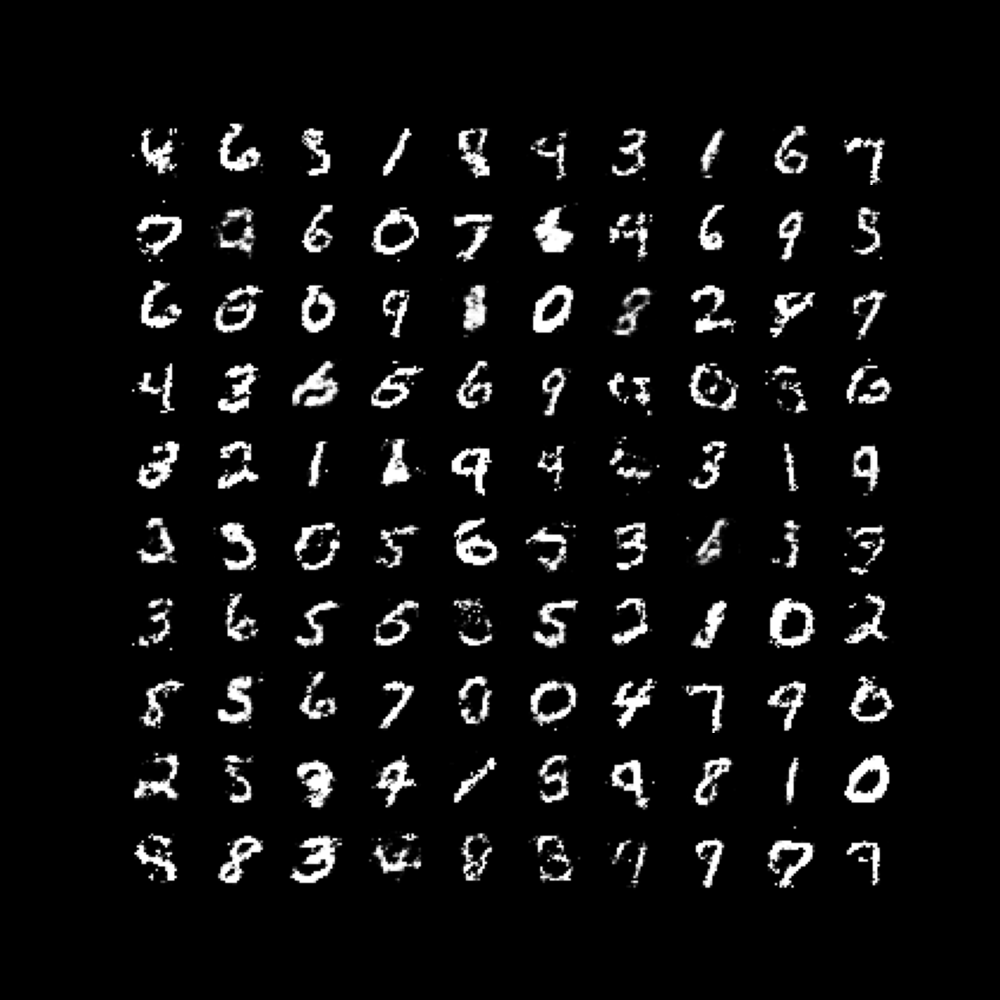}
    }\
    \subfloat[$(D_g, W_g)=(1, 50)$, $(D_f, W_f)=(3, 50)$;\\ \centering $\mathbf{Wass}_1=8.62$.] {
    \label{vis2:a}     
    \includegraphics[width=0.3\textwidth]{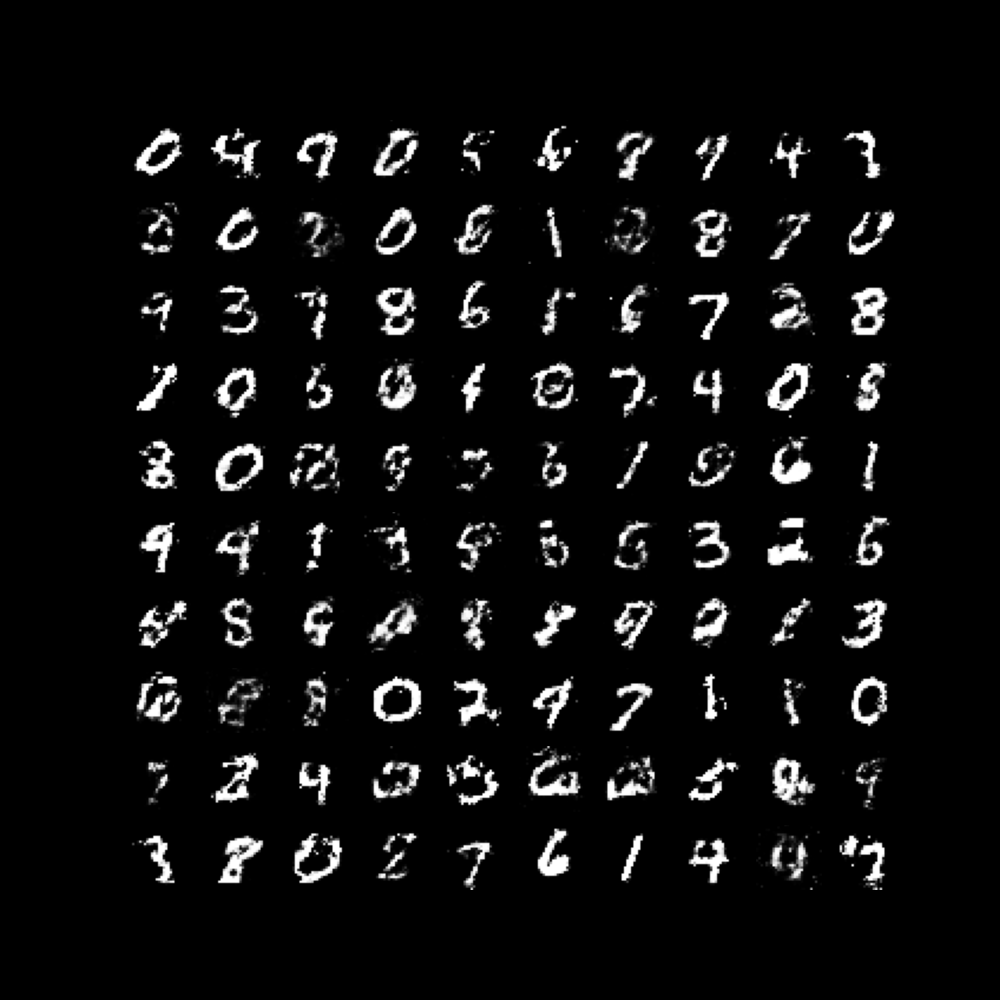}
    }\
    \subfloat[$(D_g, W_g)=(5, 50)$, $(D_f, W_f)=(3, 50)$;\\ \centering $\mathbf{Wass}_1=10.85$.] {
    \label{vis2:b}     
    \includegraphics[width=0.3\textwidth]{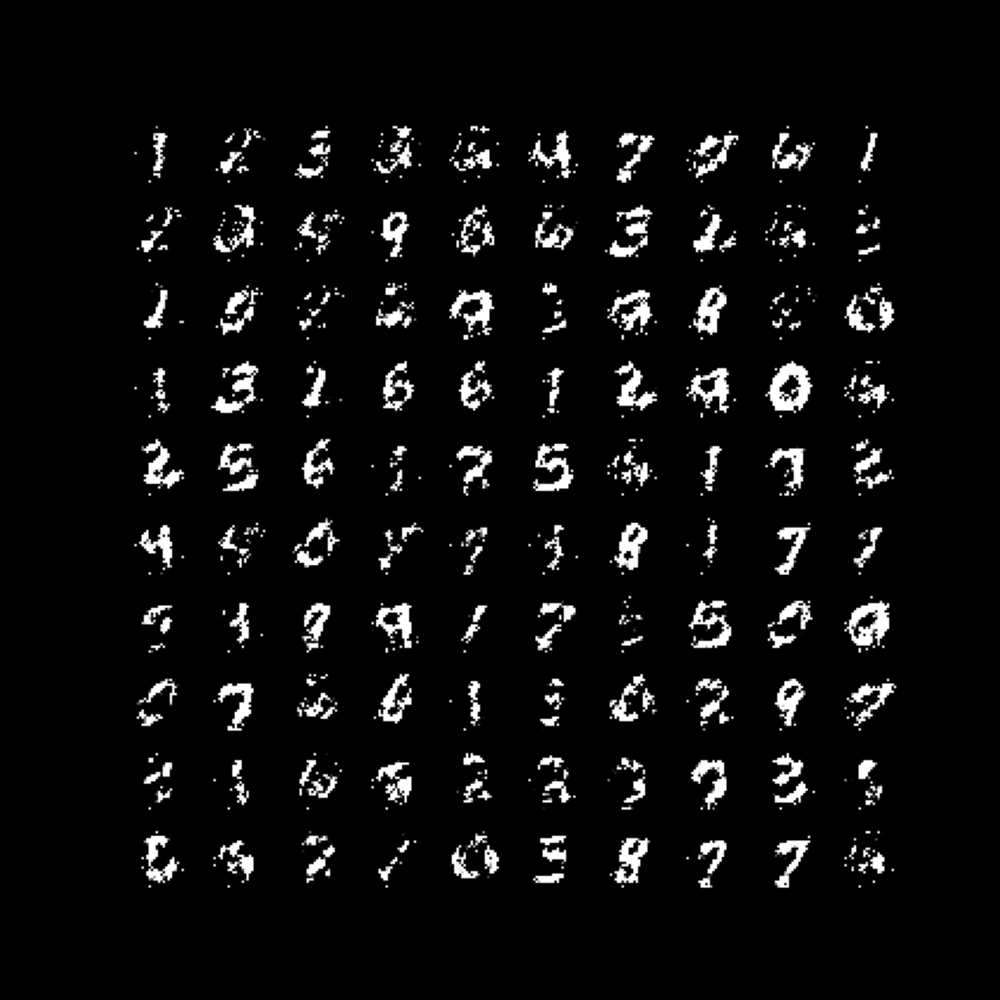}
    }\\
    \subfloat[$(D_g, W_g)=(3, 50)$, $(D_f, W_f)=(3, 150)$;\\ \centering  $\mathbf{Wass}_1=8.65$.] {
    \label{vis2:d}     
    \includegraphics[width=0.3\textwidth]{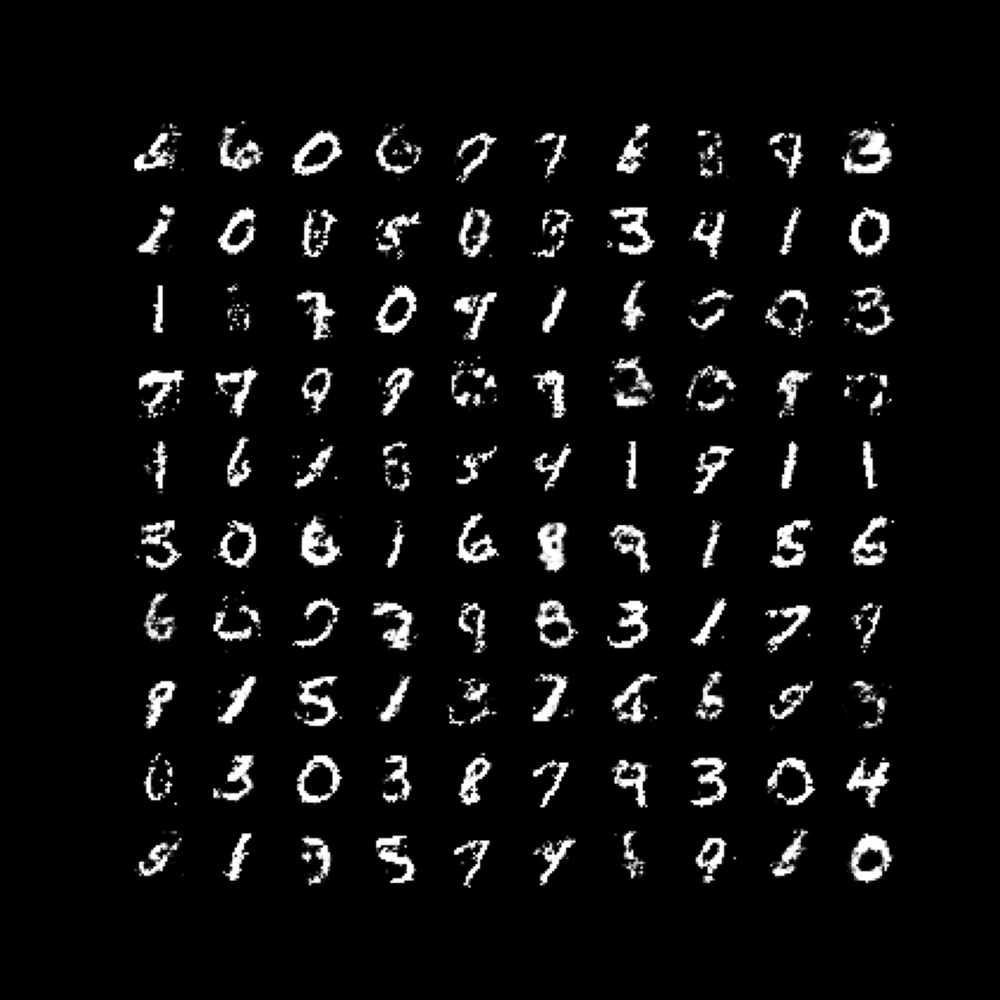}
    }\
    \subfloat[$(D_g, W_g)=(3, 50)$, $(D_f, W_f)=(10, 70)$;\\ \centering  $\mathbf{Wass}_1=8.33$.] {
    \label{vis2:e}     
    \includegraphics[width=0.3\textwidth]{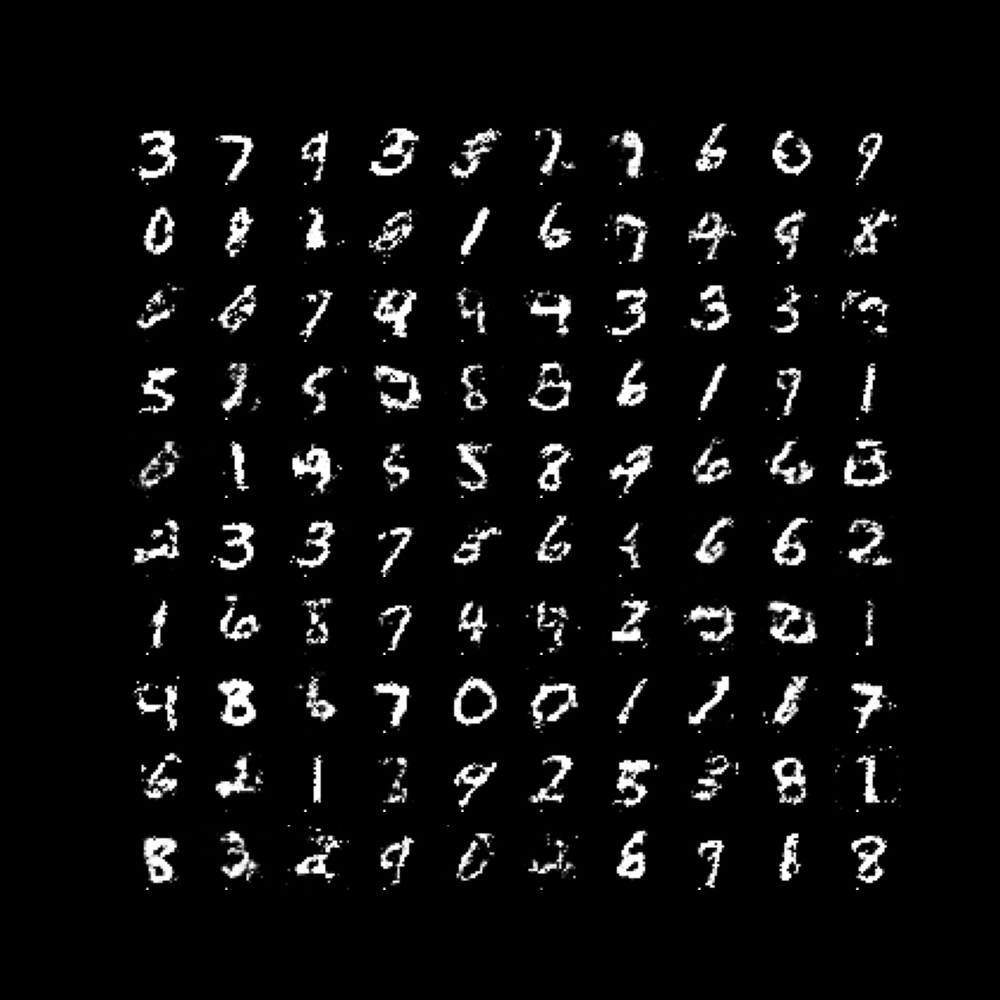}
    }\
    \subfloat[$(D_g, W_g)=(3, 200)$, $(D_f, W_f)=(5, 300)$;\\ \centering $\mathbf{Wass}_1=7.58$.] {
    \label{vis2:f}     
    \includegraphics[width=0.3\textwidth]{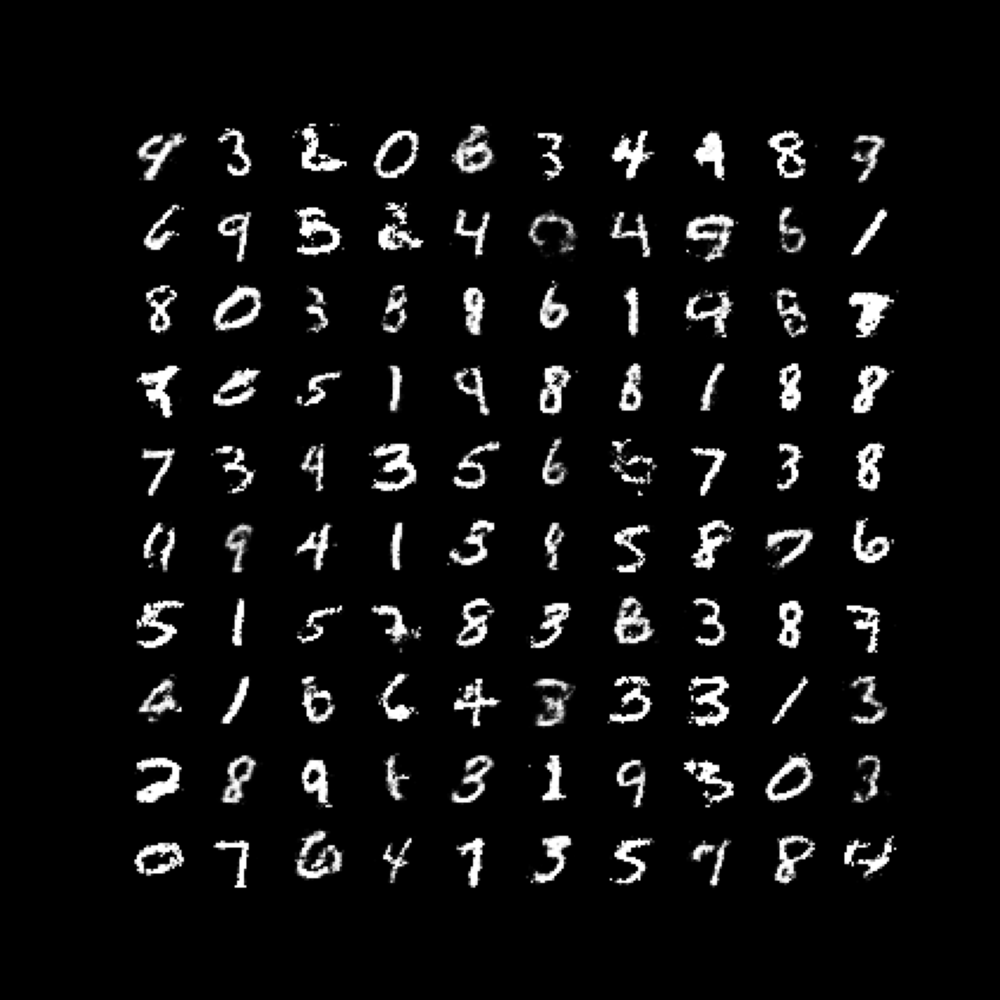}
    }\
     \caption{Visualized results.}
    \label{vis1}
\end{figure}

\cref{vis1} shows some generated data (trained on MNIST data) with different experimental settings.

\section*{Acknowledgments}
We would like to thank all editors and reviewers for their time and  constructive comments toward improving this paper. 

\bibliographystyle{siamplain}

%\bibliography{references}

\begin{thebibliography}{10}

\bibitem{anil2019sorting}
{\sc C.~Anil, J.~Lucas, and R.~Grosse}, {\em Sorting out Lipschitz function
  approximation}, in International Conference on Machine Learning, PMLR, 2019,
  pp.~291--301.

\bibitem{anthony1999neural}
{\sc M.~Anthony, P.~L. Bartlett, P.~L. Bartlett, et~al.}, {\em Neural Network
  Learning: Theoretical Foundations}, vol.~9, Cambridge University Press, 1999.

\bibitem{arjovsky2017wasserstein}
{\sc M.~Arjovsky, S.~Chintala, and L.~Bottou}, {\em Wasserstein generative
  adversarial networks}, in International Conference on Machine Learning, PMLR,
  2017, pp.~214--223.

\bibitem{arora2017generalization}
{\sc S.~Arora, R.~Ge, Y.~Liang, T.~Ma, and Y.~Zhang}, {\em Generalization and
  equilibrium in generative adversarial nets (GANs)}, in International
  Conference on Machine Learning, PMLR, 2017, pp.~224--232.

\bibitem{bai2018approximability}
{\sc Y.~Bai, T.~Ma, and A.~Risteski}, {\em Approximability of discriminators
  implies diversity in {GAN}s}, in International Conference on Learning
  Representations, 2019.

\bibitem{bernton2019parameter}
{\sc E.~Bernton, P.~E. Jacob, M.~Gerber, and C.~P. Robert}, {\em On parameter
  estimation with the Wasserstein distance}, Information and Inference: A
  Journal of the IMA, 8 (2019), pp.~657--676.

\bibitem{bjorck1971iterative}
{\sc {\AA}.~Bj{\"o}rck and C.~Bowie}, {\em An iterative algorithm for computing
  the best estimate of an orthogonal matrix}, SIAM Journal on Numerical
  Analysis, 8 (1971), pp.~358--364.

\bibitem{bowman2015generating}
{\sc S.~R. Bowman, L.~Vilnis, O.~Vinyals, A.~Dai, R.~Jozefowicz, and
  S.~Bengio}, {\em Generating sentences from a continuous space}, in
  Proceedings of The 20th {SIGNLL} Conference on Computational Natural Language
  Learning, Berlin, Germany, Aug. 2016, Association for Computational
  Linguistics, pp.~10--21.

\bibitem{brock2016neural}
{\sc A.~Brock, T.~Lim, J.~M. Ritchie, and N.~Weston}, {\em Neural photo editing
  with introspective adversarial networks}, International Conference on
  Learning Representations,  (2016).

\bibitem{cybenko1989approximation}
{\sc G.~Cybenko}, {\em Approximation by superpositions of a sigmoidal
  function}, Mathematics of Control, Signals and Systems, 2 (1989),
  pp.~303--314.

\bibitem{del2019central}
{\sc E.~Del~Barrio, J.-M. Loubes, et~al.}, {\em Central limit theorems for
  empirical transportation cost in general dimension}, The Annals of
  Probability, 47 (2019), pp.~926--951.

\bibitem{erichson2021lipschitz}
{\sc N.~B. Erichson, O.~Azencot, A.~Queiruga, L.~Hodgkinson, and M.~W.
  Mahoney}, {\em Lipschitz recurrent neural networks}, in International
  Conference on Learning Representations, 2021.

\bibitem{flamary2021pot}
{\sc R.~Flamary, N.~Courty, A.~Gramfort, M.~Z. Alaya, A.~Boisbunon, S.~Chambon,
  L.~Chapel, A.~Corenflos, K.~Fatras, N.~Fournier, L.~Gautheron, N.~T. Gayraud,
  H.~Janati, A.~Rakotomamonjy, I.~Redko, A.~Rolet, A.~Schutz, V.~Seguy, D.~J.
  Sutherland, R.~Tavenard, A.~Tong, and T.~Vayer}, {\em POT: Python optimal
  transport}, Journal of Machine Learning Research, 22 (2021), pp.~1--8.

\bibitem{glorot2010understanding}
{\sc X.~Glorot and Y.~Bengio}, {\em Understanding the difficulty of training
  deep feedforward neural networks}, in Proceedings of the Thirteenth
  International Conference on Artificial Intelligence and Statistics, JMLR
  Workshop and Conference Proceedings, 2010, pp.~249--256.

\bibitem{goodfellow2014generative}
{\sc I.~Goodfellow, J.~Pouget-Abadie, M.~Mirza, B.~Xu, D.~Warde-Farley,
  S.~Ozair, A.~Courville, and Y.~Bengio}, {\em Generative adversarial nets}, in
  Advances in Neural Information Processing Systems, vol.~27, Curran
  Associates, Inc., 2014.

\bibitem{hornik1991approximation}
{\sc K.~Hornik}, {\em Approximation capabilities of multilayer feedforward
  networks}, Neural Networks, 4 (1991), pp.~251--257.

\bibitem{huster2018limitations}
{\sc T.~Huster, C.-Y.~J. Chiang, and R.~Chadha}, {\em Limitations of the
  Lipschitz constant as a defense against adversarial examples}, in Joint
  European Conference on Machine Learning and Knowledge Discovery in Databases,
  Springer, 2018, pp.~16--29.

\bibitem{huster2021pareto}
{\sc T.~Huster, J.~Cohen, Z.~Lin, K.~Chan, C.~Kamhoua, N.~O. Leslie, C.-Y.~J.
  Chiang, and V.~Sekar}, {\em Pareto GAN: Extending the representational power
  of gans to heavy-tailed distributions}, in International Conference on
  Machine Learning, PMLR, 2021, pp.~4523--4532.

\bibitem{leskovec2009community}
{\sc J.~Leskovec, K.~J. Lang, A.~Dasgupta, and M.~W. Mahoney}, {\em Community
  structure in large networks: Natural cluster sizes and the absence of large
  well-defined clusters}, Internet Mathematics, 6 (2009), pp.~29--123.

\bibitem{liang2021well}
{\sc T.~Liang}, {\em How well generative adversarial networks learn
  distributions}, Journal of Machine Learning Research, 22 (2021), pp.~1--41.

\bibitem{NEURIPS2020_2000f632}
{\sc Y.~Lu and J.~Lu}, {\em A universal approximation theorem of deep neural
  networks for expressing probability distributions}, in Advances in Neural
  Information Processing Systems, vol.~33, Curran Associates, Inc., 2020,
  pp.~3094--3105.

\bibitem{luo2020two}
{\sc T.~Luo and H.~Yang}, {\em Two-layer neural networks for partial
  differential equations: Optimization and generalization theory}, arXiv
  preprint arXiv:2006.15733,  (2020).

\bibitem{ma2021lecturenotes}
{\sc T.~Ma}, {\em Lecture Notes for Machine Learning Theory}, 2021.

\bibitem{odena2017conditional}
{\sc A.~Odena, C.~Olah, and J.~Shlens}, {\em Conditional image synthesis with
  auxiliary classifier GANs}, in International Conference on Machine Learning,
  PMLR, 2017, pp.~2642--2651.

\bibitem{pauli2021training}
{\sc P.~Pauli, A.~Koch, J.~Berberich, P.~Kohler, and F.~Allg{\"o}wer}, {\em
  Training robust neural networks using Lipschitz bounds}, IEEE Control Systems
  Letters, 6 (2021), pp.~121--126.

\bibitem{tanielian2021approximating}
{\sc U.~Tanielian and G.~Biau}, {\em Approximating Lipschitz continuous
  functions with GroupSort neural networks}, in International Conference on
  Artificial Intelligence and Statistics, PMLR, 2021, pp.~442--450.

\bibitem{villani2008optimal}
{\sc C.~Villani}, {\em Optimal Transport: Old and New}, vol.~338 of Grundlehren
  der mathematischen Wissenschaften, Springer Berlin Heidelberg, Berlin,
  Heidelberg, 2008.

\bibitem{yang2022capacity}
{\sc Y.~Yang, Z.~Li, and Y.~Wang}, {\em On the capacity of deep generative
  networks for approximating distributions}, Neural Networks, 145 (2022),
  pp.~144--154.

\bibitem{yarotsky2017error}
{\sc D.~Yarotsky}, {\em Error bounds for approximations with deep ReLU
  networks}, Neural Networks, 94 (2017), pp.~103--114.

\bibitem{yarotsky2018optimal}
{\sc D.~Yarotsky}, {\em Optimal approximation of continuous functions by very
  deep ReLU networks}, in Conference on Learning Theory, PMLR, 2018,
  pp.~639--649.

\bibitem{zhang2022rethinking}
{\sc B.~Zhang, D.~Jiang, D.~He, and L.~Wang},  {\em Rethinking Lipschitz neural networks and certified robustness: A boolean function perspective}, in Advances in Neural
  Information Processing Systems, vol.~35, Curran Associates, Inc., 2022,
  pp.~19398--19413.
  
\end{thebibliography}

\end{document}